\documentclass[times,authoryear]{elsarticle}

\usepackage{makecell}
\usepackage{framed,multirow}

\usepackage{amssymb}
\usepackage{latexsym}
\usepackage{amsmath}
\usepackage{subcaption}
\usepackage{bm} 

\usepackage[switch]{lineno}
\usepackage{float}

\usepackage{url}
\usepackage{xcolor}
\definecolor{newcolor}{rgb}{.8,.349,.1}

\usepackage[citebordercolor=white]{hyperref}

\providecommand{\snm}[1]{#1}

\begin{document}


\begin{frontmatter}

\title{Unified Orbit-Attitude Estimation and Sensor Tasking Framework for Autonomous Cislunar Space Domain Awareness Using Multiplicative Unscented Kalman Filter}%

\author[1]{Smriti Nandan \snm{Paul}}\corref{cor1}
\author[2]{Siwei \snm{Fan}}

\affiliation[1]{organization={Department of Mechanical and Aerospace Engineering},
                addressline={Missouri University of Science and Technology},
                city={Rolla},
                postcode={65409},
                state = {MO},
                country={United States}}

\affiliation[2]{organization={Embry-Riddle Aeronautical University},
                addressline={3700 Willow Creek Rd},
                city={Prescott},
                postcode={86301},
                state = {AZ},
                country={United States}}


\begin{abstract}
Because of the strategic interests in long-term human presence on the Moon, along with increasing participation by a diverse set of space actors and easier access to space, the population of resident space objects in the cislunar regime is expected to increase over the coming decades and beyond. This anticipated traffic increase underscores the need to adapt and extend near-Earth space domain awareness (SDA) capabilities to support monitoring and characterization across the cislunar regime. In the context of SDA, the cislunar regime departs from near-Earth orbital behavior through strongly non-linear, non-Keplerian dynamics, which adversely affect the accuracy of uncertainty propagation and orbit/attitude state estimation. Additional challenges arise from long-range observation requirements, restrictive sensor-target geometry and illumination conditions, the need to monitor an expansive cislunar volume, and the large design space associated with space/ground-based sensor placement. In response to these challenges, this work introduces an advanced framework for cislunar SDA encompassing two key tasks: (1) observer architecture optimization based on a realistic cost formulation that captures key performance trade-offs, solved using the Tree of Parzen Estimators algorithm, and (2) leveraging the resulting observer architecture, a mutual information-driven sensor tasking optimization is performed at discrete tasking intervals, while orbital and attitude state estimation is carried out at a finer temporal resolution between successive tasking updates using an error-state multiplicative unscented Kalman filter. Numerical simulations demonstrate that our approach in Task 1 yields observer architectures that achieve significantly lower values of the proposed cost function than baseline random-search solutions, while using fewer sensors. Task 2 results show that translational state estimation remains satisfactory over a wide range of target-to-observer count ratios, whereas attitude estimation is significantly more sensitive to target-to-observer ratios and tasking intervals, with increased rotational-state divergence observed for high target counts and infrequent tasking updates. These results highlight important trade-offs between sensing resources, tasking cadence, and achievable state estimation performance that influence the scalability of autonomous cislunar SDA.
\end{abstract}


\end{frontmatter}


\section{Introduction}
\label{sec1}
As attention shifts toward operations beyond near-Earth orbits, safety and sustainability concerns now include the cislunar environment in addition to traditional orbital regimes. However, despite the expected rise in cislunar missions, the associated space domain awareness (SDA), space situational awareness (SSA), and space traffic management (STM) frameworks remain significantly less mature than those supporting near-Earth operations. The anticipated growth in cislunar activity is driven not only by the Artemis program and evolving exploration architectures \citep{9172323, NASA_M2M_ADD_2024}, but also by coordinated international (cis)lunar exploration efforts led by agencies such as ESA, JAXA, ISRO, and CNSA \citep{ISECG_GER_2024}, together with increasing commercial lunar missions \citep{lederer2023commercial} whose transit and support requirements place additional operational demands on the cislunar domain.\\

Cislunar SDA is substantially more challenging than its near-Earth counterpart due to a combination of observational limitations and complex dynamical effects. Long observation ranges across the expansive cislunar domain yield degraded signal-to-noise ratios and restrictive observation geometries. These challenges are further compounded by limited sensing assets, line-of-sight occultations, eclipse-induced illumination losses, and Sun-Earth-Moon exclusion-zone constraints, which significantly reduce viable observation opportunities. Moreover, the strongly non-linear, non-Keplerian dynamics governing the cislunar regime lead to increased uncertainty in state propagation and adversely affect observability. The absence of established design heuristics for cislunar observer architectures, combined with competing requirements, necessitates systematic optimization of observer placement and observation resource allocation to enable an effective and scalable cislunar SDA framework.\\

Because ground-based telescopic observation of cislunar objects is limited by unfavorable and time-varying viewing geometries, atmospheric interference, restricted illumination windows, and severe signal degradation at long observation ranges, this study focuses on passive electro-optical sensing from space-based platforms deployed in cislunar orbits. Within this space-based observation context, the study focuses on two related tasks motivated by cislunar SDA objectives (specific SDA requirements introduced later in the paper): (1) a Bayesian optimization approach is used to identify effective observer architectures under cislunar dynamics represented by a simplified circular restricted three-body problem (CR3BP) model; and (2) building on these architectures, a coordinated multi-sensor tasking and estimation strategy is developed that enables orbital and angular state estimation at a temporal resolution higher than the tasking updates.\\

Task 1 defines the feasible observer trajectories considered for space-based sensing by selecting candidate orbits from several families of periodic solutions of the CR3BP. These orbit families span a range of dynamical behaviors and provide distinct viewing geometries and coverage characteristics relevant to cislunar monitoring. The set of objects to be observed during Task 1 (described later in the paper) is modeled as stationary in the CR3BP rotating frame. Their locations are generated through a selection process informed by existing and anticipated cislunar missions \citep{johnson2022fly}. Together, these target locations span a broad region of the cislunar domain and are intended to represent a diverse range of plausible future mission scenarios. A novel cost function is developed to capture key performance trade-offs for optimizing multi-spacecraft observer architectures. Pragmatic, real-world constraints are embedded directly within the objective formulation to maintain operational relevance. The optimization problem is solved using a Bayesian optimization approach designed for complex, high-dimensional design spaces, offering efficient exploration of potentially discontinuous search domains relevant to cislunar applications.\\     

Task 2 in this study involves the development of a sensor tasking strategy integrated with state estimation. Target orbits are drawn from representative cislunar orbit families and mission-relevant trajectories, with optical sensor parameters specified to reflect realistic monitoring scenarios. With a large number of observers and candidate targets, multi-step sensor tasking becomes computationally intractable due to the combinatorial growth of possible assignments. Therefore, sensor tasking is performed using a greedy, single-step strategy that maximizes a mutual information-based objective at each tasking epoch \citep{doi:10.2514/6.2024-1676}, assuming cooperative sensing among observers. Sensor tasking decisions are updated at a coarser temporal cadence, while the target's orbital position and velocity, attitude, and angular velocity states are estimated at a finer time resolution using an error-state multiplicative unscented Kalman filter (UKF) suitable for nonlinear state propagation and measurement models (brightness, right ascension, declination). Attitude errors are parameterized using generalized Rodrigues parameters. This structure supports numerical tests that vary tasking intervals and observer-to-target ratios to assess estimation performance.\\ 

The realism of simulated optical measurements in the cislunar environment is influenced by the surface reflectance model used to represent target brightness. Optical brightness is commonly formulated through the bidirectional reflectance distribution function (BRDF), with space objects represented as collections of small planar facets to capture geometry/material-dependent reflectance effects. In this work, brightness is computed using the physically based Cook-Torrance BRDF formulation \citep{cook1982reflectance, wetterer2009attitude}. The model captures microfacet-based specular and diffuse reflection effects, including surface roughness, Fresnel behavior, and geometric shadowing and masking under varying illumination and viewing geometries. Prior investigations have demonstrated the applicability of physically based BRDF models, including Cook-Torrance and Ashikhmin-Shirley, for modeling brightness of space objects \citep{wetterer2009attitude, linares2014space}. Empirical BRDF models (e.g., Neumann and Strauss \citep{montes2012overview}), which are not derived from first-principles reflectance theory and may not strictly enforce physical constraints such as energy conservation or reciprocity, are not considered in this study. Finally, to reflect realistic sensing conditions, appropriate measurement noise is applied to the ideal simulated brightness signals.\\

Sun's position is obtained from high-fidelity SPICE (Spacecraft, Planet, Instrument, C-matrix, Events) ephemerides \citep{SPICE} and transformed into the CR3BP rotating frame at each time step to provide realistic illumination geometry. Observer and target trajectories, however, are propagated using the CR3BP equations of motion rather than full ephemeris models. This approach maintains realistic lighting conditions while simplifying the dynamical model for computational efficiency.\\

This paper is organized as follows. Section~\ref{sec:back} summarizes the modeling and estimation background, including the CR3BP dynamical framework, the optical brightness modeling framework, and the error-state multiplicative UKF employed for orbit-attitude estimation. Section~\ref{sec:arch_opt} describes the observer architecture optimization framework (Task 1), covering the selection of candidate observer orbits, target representations, and the Bayesian optimization approach used to explore the design space. Section~\ref{sec:sensor_Tk} presents the integrated sensor tasking and state estimation methodology (Task 2) developed on top of the optimized architectures. Numerical simulation results are presented and discussed in Section~\ref{sec:results}. Section~\ref{sec:conclusions} concludes the paper and outlines directions for future work.

\section{Modeling and Estimation Background}\label{sec:back}
\subsection{CR3BP Orbital Dynamics}
The orbital motion considered in this work is modeled using the CR3BP. In this formulation, two massive primary bodies, denoted $P_1$ and $P_2$ with respective mass $m_1$ and $m_2$, move on circular orbits about their common barycenter. A third body, $P_3$, is assumed to have negligible mass and therefore does not influence the motion of the primaries. The motion of $P_3$ is governed solely by the gravitational attraction of $P_1$ and $P_2$. Within this framework, the trajectories of the sensing platforms and the target objects are propagated independently under the CR3BP assumptions.\\

The CR3BP is commonly formulated in a rotating reference frame in which the two primary bodies remain fixed. In this frame, the $x$-axis is defined along the line connecting the primaries, oriented from $P_1$ toward $P_2$. The $z$-axis is parallel to the angular momentum vector associated with the circular motion of the primaries and is normal to their orbital plane. The $y$-axis completes the right-handed coordinate system. The origin of the rotating frame is at the barycenter of $P_1$ and $P_2$. A dimensionless mass parameter, $\mu$, specifies the location of the barycenter along the $x$-axis and is defined as the ratio of the mass of $P_2$ to the total system mass. Smaller values of $\mu$ correspond to systems where the barycenter lies closer to $P_1$.\\

For the Earth-Moon system, the Earth is designated as the primary body $P_1$ and the Moon as the second primary body $P_2$. An important quantity in CR3BP is the mass parameter $\mu$, or the mass ratio of the two primaries. A value of $\mu=0.01215$ is adopted for the Earth-Moon system. Furthermore, to improve numerical conditioning, the governing equations of motion are often expressed in non-dimensional form, where lengths are normalized by the Earth-Moon separation $l^*$. As a result, the Earth-Moon distance is unity in the normalized form. Time is also non-dimensionalized using the characteristic time scale $t^*=1/n$, where $n$ is Moon's mean motion. Under this normalization, the Earth and the Moon remain fixed along the $x$-axis at $\overline{r}_{P_1}=[-\mu,0,0]^T$ and $\overline{r}_{P_2}=[1-\mu,0,0]^T$, respectively. All state variables for the third body are defined relative to the Earth-Moon barycenter and this rotating frame. Conversion back to physical units is performed by scaling non-dimensional distances by $l^*$ and time-dependent quantities by $t^*$.\\

The translational motion of the third body $P_3$ is described by the six-dimensional state vector $\textbf{x}=[x,y,z,\dot{x},\dot{y},\dot{z}]$, which evolves according to the CR3BP dynamics in the rotating frame. The dynamics are governed by a coupled set of nonlinear second-order differential equations that include the gravitational attraction of both primaries as well as Coriolis and centrifugal effects from the rotating coordinate system:
\begin{align}
\ddot{x} - 2\dot{y} &= x
- (1-\mu)\frac{x+\mu}{d^3}
- \mu\frac{x-1+\mu}{r^3} \\
\ddot{y} + 2\dot{x} &= y
- \left[(1-\mu)\frac{1}{d^3} + \mu\frac{1}{r^3}\right] y \\
\ddot{z} &= -\left[(1-\mu)\frac{1}{d^3} + \mu\frac{1}{r^3}\right] z
\end{align}

where $d$ and $r$ denote the distances from $P_3$ to the primary body $P_1$ and the second primary body $P_2$, respectively. They are defined as:
\begin{align}
d = \sqrt{y^2 + z^2 + (x+\mu)^2}, \qquad
r = \sqrt{y^2 + z^2 + (x-1+\mu)^2}
\end{align}

The right-hand sides of the CR3BP equations of motion can also be expressed as partial derivatives of a pseudo-potential function $U(x,y,z)$ with respect to the spatial coordinates. This pseudo-potential combines the gravitational contributions of both primaries with the centrifugal potential from the rotating frame:

\begin{equation}
    U = U_c + U_g, \qquad
U_c = \frac{1}{2}(x^2+y^2), \qquad
U_g = \frac{1-\mu}{d} + \frac{\mu}{r}
\end{equation}

The pseudo-potential $U(x,y,z)$ is directly related to the Jacobi constant ($JC$) which is the only conserved quantity in the CR3BP. This invariant is defined as $\mathrm{JC} = 2U - v^2$, where $v = \sqrt{\dot{x}^2 + \dot{y}^2 + \dot{z}^2}$ is the speed of $P_3$ in the rotating frame. Since the Jacobi constant is conserved along any CR3BP trajectories, it provides a numerical check for orbit propagation accuracy. More importantly, for a fixed value of the Jacobi constant, the configuration space is partitioned into dynamically accessible and forbidden regions. In regions where $U < \mathrm{JC}/2$, the velocity squared $v^2 = 2U - \mathrm{JC}$ becomes negative, making those regions physically inaccessible to $P_3$. The boundaries separating the allowed and forbidden regions are referred to as zero-velocity curves in the planar case or zero-velocity surfaces in three dimensions \citep{koon_dynamical_2000}. The topology of these surfaces varies with the Jacobi constant values and governs global transport mechanisms and connectivity between regions of the Earth-Moon system.

\subsection{Optical Brightness Modeling}\label{chap:brightness}
\begin{figure}[!ht]
  \begin{center}
  \includegraphics[width=22em]{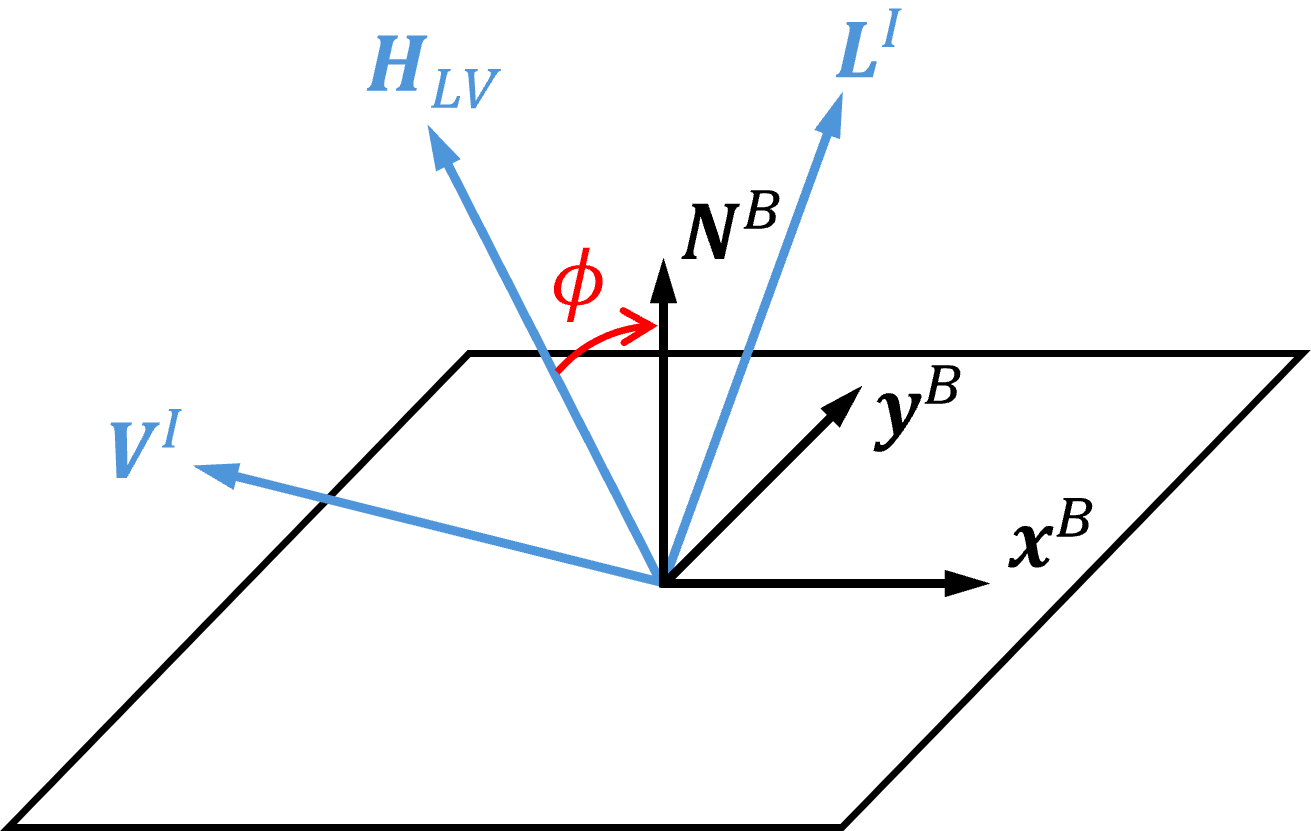}
  \caption{Reflection Geometry for a Representative Surface Facet. (quadrilateral facet shown for illustration purpose, triangular facet is used.)}
  \label{fig:facet_geometry}
  \end{center}
\end{figure}

For space-based electro-optical sensing, object brightness is characterized using apparent magnitude, which expresses observed intensity relative to the Sun. The solar reference magnitude $m_{\odot,\Delta\lambda}$ corresponds to the Sun's apparent magnitude in the sensor's effective bandpass $\Delta\lambda$ ($m_{\odot,V}\approx -26.74$ for Johnson V-band system). The apparent magnitude of a space object is given by:
\begin{equation}\label{eq:rel_mag}
m_{\mathrm{obj}}
= m_{\odot,\Delta\lambda} - 2.5\left[\log_{10}(F_t) - \log_{10}(I_{\odot,\Delta\lambda})\right]
\end{equation}
where $I_{\odot,\Delta\lambda}$ is the solar irradiance at 1 AU integrated over the sensor's bandpass and $F_t$ is the band-limited irradiance the sensor receives from the object. The measured irradiance $F_t$ consists of the reflected optical flux from the object surface combined with measurement noise from the sensing hardware. We model measurement uncertainty solely through charge-coupled device (CCD) noise:
\begin{equation}
F_t = V_{\mathrm{CCD}} + \sum_{i = 1}^N F_V(i)
\end{equation}
where $F_V(i)$ is the irradiance from the $i^{\text{th}}$ facet, $N$ is the total number of facets, and $V_{\mathrm{CCD}}$ is additive noise in irradiance-equivalent units. The object is discretized into small planar facets to capture geometry/attitude-dependent photometric effects, with each facet contributing according to its orientation, illumination geometry, and surface reflectance. Figure~\ref{fig:facet_geometry} shows the geometric quantities needed to evaluate the photometric contribution of an individual surface facet.
In Fig.~\ref{fig:facet_geometry}, $\textbf{L}^I$ and $\textbf{V}^I$ denote unit vectors pointing from the facet toward the Sun and the observing space-based optical sensor, respectively. The half-vector $\textbf{H}_{LV}$ is defined as the normalized bisector of the illumination and viewing direction vectors. Each facet has a local body-fixed orthonormal frame $\{\textbf{x}^B,\textbf{y}^B,\textbf{N}^B\}$, where $\textbf{N}^B$ is the outward unit normal defining the facet orientation and $\textbf{x}^B,\textbf{y}^B$ span the facet plane. The angle $\phi$ is measured between the facet normal and the half-vector. Note that in practice, the triangular facet (defined with 3 vertices) is used to guarantee local co-planarity.\\ 

Since the Sun and observer are far from the object compared to its size, facet positions are approximated by the object's center (its orbital location) when computing viewing directions and illumination conditions. The Sun’s inertial position is obtained from ephemerides provided by the NASA Jet Propulsion Laboratory (JPL) NAIF SPICE system in the heliocentric J2000 frame and is transformed into the CR3BP rotating frame at each time step to ensure consistency with the dynamical model.\\

The irradiance at the observer from the $i^{th}$ surface facet, $F_V(i)$, is evaluated as: 
\begin{equation}\label{eq:cdcs}
F_V(i)
= \frac{I_{\odot,\Delta\lambda}}{4\pi r^2}\,\mathcal{R}_i
\quad \textrm{where} \quad
\mathcal{R}_i
= A(i)\,\rho_t(i)\,
\big(\mathbf{N}^B(i)\cdot\mathbf{L}^I\big)\,
\big(\mathbf{N}^B(i)\cdot\mathbf{V}^I\big)
\end{equation}
where $A(i)$ denote the area of the $i^{th}$ facet and $r$ is the distance between the observer and the space object. The dot products represent foreshortening in illumination and viewing geometry, while the $\frac{1}{4\pi r^2}$ term captures inverse-square spreading of reflected flux. The total reflectance $\rho_t(i)$ depends on the adopted surface reflection model. When a BRDF formulation is used, $\rho_t(i)$ is typically expressed as a weighted combination of diffuse and specular components \citep{wetterer_refining_2014}:

\begin{equation}\label{eq:rhods}
\rho_t(i) = \sum_{k \in \{d,s\}} w_k\,\rho_k(i)
\quad \textrm{where} \quad
w_d = d,\; w_s = s
\end{equation}
The specific functional forms of $\rho_{d}(i)$ and $\rho_{s}(i)$ depend on the chosen BRDF model. In general, BRDF models may be categorized as empirical, theoretical, experimental, or hybrid approaches, depending on how surface scattering behavior is represented \citep{wetterer2014refining,montes2012overview}. We adopt the Cook-Torrance \citep{cook1982reflectance} BRDF model, which captures specular reflection using microfacet model at modest computational cost.\\

The Cook-Torrance reflectance model describes surface reflection using a microfacet formulation, where each macroscopic facet represents an ensemble of unresolved microscopic surface elements \citep{cook1982reflectance}. Instead of assigning a single albedo to a facet, its reflectance is expressed through statistical functions that represent the collective behavior of these microfacets. The model is defined by three terms: the Beckmann microfacet slope distribution $D_B$ \citep{beckmann1987scattering}, a geometric attenuation factor $\alpha$ accounting for shadowing and masking on the $i^{\textrm{th}}$ facet, and a Fresnel term $F$ describing light reflectance:
\begin{align}
D_B
&=
\delta_{RMS}^{-2}\,
\cos^{-4}\!\phi\,
\exp\!\left(-\frac{\tan^2\!\phi}{\delta_{RMS}^2}\right)
\\[6pt]
\alpha(i)
&=
\operatorname{min}\!\Bigg(
1,\;
\frac{2\big(\mathbf{N}^B(i)\!\cdot\!\mathbf{H}_{LV}\big)
      \big(\mathbf{N}^B(i)\!\cdot\!\mathbf{V}^I\big)}
     {\mathbf{H}_{LV}\!\cdot\!\mathbf{V}^I},\;
\frac{2\big(\mathbf{N}^B(i)\!\cdot\!\mathbf{H}_{LV}\big)
      \big(\mathbf{N}^B(i)\!\cdot\!\mathbf{L}^I\big)}
     {\mathbf{H}_{LV}\!\cdot\!\mathbf{V}^I}
\Bigg)
\\[6pt]
F
&=
\frac{(p-q)^2}{2(p+q)^2}
\left[
1 +
\left(
\frac{q(p+q)-1}{q(p-q)+1}
\right)^2
\right]
\end{align}


The parameter $\delta_{RMS}$ represents the root mean square slope of the microfacet population and describes surface roughness; it is either specified as a material constant, derived from empirical roughness models, or estimated by fitting to photometric observations \citep{cook1982reflectance,bagher_rms_2012,matusik_reflectance_2003,beckmann1987scattering}. The Fresnel term $F$ is evaluated using the auxiliary variables $p$ and $q$, defined as: 
\begin{align}
q &= \textbf{V}^I\cdot \textbf{H}_{LV}\\
p &= (q^2 - 1) + b^2\\
b &= \frac{1+\sqrt{F_0}}{1-\sqrt{F_0}}
\end{align}  
where $F_0$ represents the surface reflectance under normal illumination. Using these definitions, the specular contribution from the $i^{\text{th}}$ facet is expressed as:
\begin{equation}
\rho_s(i)
=
\frac{\alpha(i)\,D_B\,F}{\pi}
\left[
(\mathbf{N}^B(i)\!\cdot\!\mathbf{L}^I)
(\mathbf{N}^B(i)\!\cdot\!\mathbf{V}^I)
\right]^{-1}
\end{equation}
The diffuse component in the Cook-Torrance model follows a Lambertian formulation and is independent of viewing direction:
\begin{equation}
\rho_d(i)
=
\frac{1}{\pi}\,\rho(i)
\end{equation}
where $\rho(i)$ is the diffuse albedo  of the $i^{\text{th}}$ facet. With $\rho_d(i)$ and $\rho_s(i)$ defined, the total reflected contribution is obtained from Eq.~\ref{eq:rhods}.

\subsection{Analytical Brightness Models}\label{chap:analytical_brightness}
While the facet-based reflection model will be used during the estimation task, an analytical brightness model based on Lambertian, sphere-shaped targets is used in the architecture optimization task due to computational considerations. The analytical model is a function of only the solar phase angle, $\alpha$, and does not include a specular reflectance component. The following analytical brightness model is adopted in the apparent brightness magnitude equation (Eq.~\ref{eq:rel_mag}).
\begin{align}\label{eq:mag_sph}
m_{\mathrm{obj}}
&= m_{\mathrm{Sun}}
- 2.5 \log_{10}\!\left(
\frac{2 C_d R_{\mathrm{obj}}^{2}}{3 \pi \lVert \vec{r}_{OT} \rVert^{2}}
\, \Phi(\alpha)
\right) \\
\Phi(\alpha)
&= \sin \alpha + (\pi - \alpha)\cos \alpha \\
\alpha
&= \cos^{-1}\!\left(
\frac{\vec{r}_{OT}\cdot\vec{r}_{ST}}
{\lVert \vec{r}_{OT} \rVert \, \lVert \vec{r}_{ST} \rVert}
\right)
\end{align}
where $C_d$ is the surface diffuse reflection coefficient, $R_{\textrm{obj}}$ is the radius of the assumed spherical targets, $\vec{r}_{OT}$ is the observer-to-target position vector, and $\vec{r}_{ST}$ is the Sun-to-target position vector. The phase angle $\alpha$ is indicative of the target's illumination and viewing geometry, and it is computed from $\vec{r}_{OT}$ and $\vec{r}_{ST}$.

\subsection{Error-State Multiplicative Unscented Kalman Filter}
This work uses the error-state multiplicative UKF  \citep{doi:10.2514/2.5102} to estimate a space object's orbital and attitude states. The measurements consist of observer-frame line-of-sight angles (right ascension and declination) and photometric magnitude (Section~\ref{chap:brightness}). Attitude is represented globally with a quaternion (singularity-free kinematics), while attitude errors are locally parameterized with generalized Rodrigues parameters (GRPs), which gives a minimal (3-parameter) representation that integrates naturally with the UKF's sigma-point formulation. The angular measurements are most informative for the translational states, whereas photometric measurements can contribute to attitude estimation. The combined attitude-translation state vector for our estimation problem is:
\begin{align}
\hat{\mathbf{x}}_k \triangleq
\begin{bmatrix}
\delta \hat{\mathbf{p}}^\top_k &
\hat{\boldsymbol{\omega}}^\top_k &
\hat{\mathbf{r}}^\top_k &
\hat{\mathbf{v}}^\top_k
\end{bmatrix}^\top
\end{align}
where $\delta \hat{\mathbf{p}}$ is the local attitude-error (error GRP) vector associated with the global attitude quaternion estimate $\hat{\mathbf{q}}$, which represents the orientation of the body frame with respect to the inertial frame. The vector $\hat{\boldsymbol{\omega}}$ is the estimated angular velocity of the body with respect to the inertial frame, expressed in the body frame. The vectors $\hat{\mathbf{r}}$ and $\hat{\mathbf{v}}$ are the estimated position and velocity of the object expressed in the rotating frame of the CR3BP. The initial attitude-error state estimate is set to zero, i.e., $\delta \hat{\mathbf{p}}_0 = \mathbf{0}$ \citep{doi:10.2514/2.5102}.



The continuous-time system dynamics and discrete-time measurement model are described by:
\begin{align}
\dot{\mathbf{x}} &= f(\mathbf{x},t) + B(\mathbf{x},t)\,\boldsymbol{\eta}(t) \\
\mathbf{y}_k &= h(\mathbf{x}_k,t_k) + \boldsymbol{\epsilon}_k,
\qquad \boldsymbol{\epsilon}_k \sim \mathcal{N}(\mathbf{0},R_k)
\end{align}
where $\mathbf{x}$ denotes the system state, $f(\cdot)$ represents the nonlinear state dynamics, and $B(\cdot)$ is the process-noise influence matrix. The term $\boldsymbol{\eta}(t)$ denotes zero-mean white process noise. The function $h(\cdot)$ defines the nonlinear measurement model, $\boldsymbol{\epsilon}_k$ is additive Gaussian measurement noise with covariance $R_k$, and $\mathbf{y}_k$ is the measurement vector at time $t_k$.


To approximate the propagation of the state mean and covariance through the nonlinear models, the UKF employs a deterministic set of sigma points and associated weights. For a combined state dimension $n$ (with $n=12$ in this study) with state mean $\boldsymbol{\mu}_k$ and associated covariance $P_k$, the UKF represents the state distribution using a set of $2n+1$ sigma points, constructed as:
\begin{align}
\boldsymbol{\Sigma}_k &= \sqrt{(n+\lambda)\,P_k}\\
\boldsymbol{\chi}_k^{(0)} &= \boldsymbol{\mu}_k\\
\boldsymbol{\chi}_k^{(i)} &= \boldsymbol{\mu}_k + \boldsymbol{\Sigma}_k^{(:,i)}, \quad i=1,\dots,n\\
\boldsymbol{\chi}_k^{(i)} &= \boldsymbol{\mu}_k - \boldsymbol{\Sigma}_k^{(:,i-n)}, \quad i=n+1,\dots,2n
\end{align}
where $\boldsymbol{\Sigma}_k^{(:,i)}$ denotes the $i$th column of the scaled square-root covariance. Each sigma point $\boldsymbol{\chi}_k^{(i)}$ is assigned a corresponding weight that determines its contribution to the reconstructed mean and covariance after nonlinear propagation:
\begin{align}
W_0^{\mathrm{mean}} &= \frac{\lambda}{n+\lambda}\\
W_0^{\mathrm{cov}} &= \frac{\lambda}{n+\lambda} + \left(1-\alpha^2+\beta\right)\\
W_i^{\mathrm{mean}} &= W_i^{\mathrm{cov}} = \frac{1}{2(n+\lambda)}, \quad i=1,\dots,2n
\end{align}
The composite scaling parameter $\lambda$ is defined as:
\begin{equation}
\lambda = \alpha^2(n+\kappa)-n
\end{equation}
where $\alpha$ controls the overall spread of the sigma points, $\beta$ reflects prior assumptions on the state distribution, and $\kappa$ is an unscented transform tuning parameter. In this work, the parameters are chosen as $\kappa= 3 - n$, $\alpha=0.5$, and $\beta=2$.

The error-state multiplicative UKF represents attitude errors using GRPs, while the global attitude is propagated as a unit quaternion. First, for each attitude-error sigma point $\bm{\chi}_{\delta p,k}^{(i)}$, we form the corresponding error quaternion $\delta \textbf{q}_k^{(i)}$ using:
\begin{align}
\delta\bm{\varrho}_k^{(i)} &= f^{-1}\!\left(a+\delta q^{(i)}_{k,4}\right)\,\bm{\chi}_{\delta p,k}^{(i)}\\
\delta q^{(i)}_{k,4} &= \frac{-a\,||\bm{\chi}_{\delta p,k}^{(i)}||^2 + f\sqrt{f^2+(1-a^2)||\bm{\chi}_{\delta p,k}^{(i)}||^2}}{{f^2+||\bm{\chi}_{\delta p,k}^{(i)}||^2}} \\
\delta \textbf{q}_k^{(i)} &= 
\begin{bmatrix}\delta\bm{\varrho}_k^{(i)}\\[5pt] \delta q^{(i)}_{k,4}\end{bmatrix}
\end{align}
where $a \in [0,1]$ is a tuning parameter and
$f = 2(a+1)$ is a scale factor. In this work, $a = 0.5$ is adopted, yielding $f = 3$. Next, the quaternion sigma points are obtained by applying the error quaternion sigma points to the mean quaternion estimate via quaternion multiplication:
\begin{align}
\hat{\bm{q}}^{(0)}_k &= \hat{\bm{q}}^{+}_k\\
\hat{\bm{q}}^{(i)}_k &= \delta \bm{q}^{(i)}_k \otimes \hat{\bm{q}}^{+}_k,\quad i=1,2,\cdots,2n
\end{align}

The sigma points $\boldsymbol{\chi}^{(i)}$ are propagated from the current time $t_k$ to $t_{k+1}$ by integrating the CR3BP translational dynamics and the torque-free attitude dynamics:
\begin{align}
\dot{\boldsymbol{\chi}}^{(i)} = f\!\left(\boldsymbol{\chi}^{(i)},\,\hat{\boldsymbol{q}}^{(i)}\right)
\end{align}



From the propagated sigma points, the error quaternion sigma points are calculated, followed by their conversion to error GRP sigma points:
\begin{align}
\delta \hat{\mathbf{q}}_{k+1}^{-(i)}
&=
\hat{\mathbf{q}}_{k+1}^{-(i)} \otimes \left(\hat{\mathbf{q}}_{k+1}^{-(0)}\right)^{-1},
\qquad
\left(\hat{\mathbf{q}}_{k+1}^{-(0)}\right)^{-1}
=
\begin{bmatrix}
-\hat{\boldsymbol{\varrho}}_{k+1}^{-(0)}\\[3pt]
\hat{q}_{k+1,4}^{-(0)}
\end{bmatrix}
\\[6pt]
\delta \hat{\boldsymbol{p}}_{k+1}^{-(i)}
&=
\frac{f}{a+\delta \hat{q}_{k+1,4}^{-(i)}}\,
\delta \hat{\boldsymbol{\varrho}}_{k+1}^{-(i)} 
\end{align}


The error GRP corresponding to the mean sigma point is then set to zero. The \textit{a priori} state estimate and its associated error covariance at time $t_{k+1}$ are reconstructed from the propagated sigma points (error GRP, angular velocity, position and velocity) using the corresponding unscented weights, as: 
\begin{align}
\hat{\mathbf{x}}^{-}_{k+1}
&=
\sum_{i=0}^{2n}
W_i^{\mathrm{mean}}\,\boldsymbol{\chi}^{-\,(i)}_{k+1} \\
P^{-}_{k+1}
&=
\sum_{i=0}^{2n}
W_i^{\mathrm{cov}}
\left(
\boldsymbol{\chi}^{-\,(i)}_{k+1}
-
\hat{\mathbf{x}}^{-}_{k+1}
\right)
\left(
\boldsymbol{\chi}^{-\,(i)}_{k+1}
-
\hat{\mathbf{x}}^{-}_{k+1}
\right)^{\!\top}
+ Q_{k+1}
\end{align}
where, $Q_{k+1}$ denotes the process-noise covariance matrix. Using the resulting \textit{a priori} mean and covariance, a new set of sigma points is generated and transformed through the nonlinear measurement model. Each sigma point is propagated through the measurement model to generate a predicted observation
$\boldsymbol{\gamma}_{k+1}^{(i)}$, which consists of the observer-frame angular measurements
(right ascension and declination) and the predicted photometric brightness (described in
Section~\ref{chap:brightness}). The actual sensor measurement at time $t_{k+1}$ is denoted by
$\boldsymbol{y}_{k+1,meas}$.
\begin{align}
\boldsymbol{\gamma}_{k+1}^{(i)}
&=
h\!\left(
\boldsymbol{\chi}_{k+1}^{-\,(i)},
\hat{\boldsymbol{q}}_{k+1}^{-\,(i)}
\right) \\
\boldsymbol{y}_{k+1,meas}
&\triangleq
\begin{bmatrix}
m_{\mathrm{rel,meas}}, &
\mathrm{RA_{meas}}, &
\mathrm{DEC_{meas}}
\end{bmatrix}^{\!\top}
\end{align}
The predicted measurement mean is then computed as the weighted average of the predicted measurement sigma points using the unscented weights:
\begin{align}
\hat{\boldsymbol{y}}_{k+1}^{-}
&=
\sum_{i=0}^{2n}
W_i^{\mathrm{mean}}\,\boldsymbol{\gamma}_{k+1}^{(i)}
\end{align}
Next, the innovation covariance and the state-measurement cross-covariance matrices are computed as:
\begin{align}
P^{yy}_{k+1}
&=
\sum_{i=0}^{2n}
W_i^{\mathrm{cov}}
\left(
\boldsymbol{\gamma}_{k+1}^{(i)} - \hat{\boldsymbol{y}}_{k+1}^{-}
\right)
\left(
\boldsymbol{\gamma}_{k+1}^{(i)} - \hat{\boldsymbol{y}}_{k+1}^{-}
\right)^{\!\top} \\[6pt]
P^{\nu\nu}_{k+1}
&=
P^{yy}_{k+1} + R_{k+1} \\[6pt]
P^{xy}_{k+1}
&=
\sum_{i=0}^{2n}
W_i^{\mathrm{cov}}
\left(
\boldsymbol{\chi}_{k+1}^{-\,(i)} - \hat{\boldsymbol{x}}_{k+1}^{-}
\right)
\left(
\boldsymbol{\gamma}_{k+1}^{(i)} - \hat{\boldsymbol{y}}_{k+1}^{-}
\right)^{\!\top}
\end{align}
The covariance matrices further feed into the computation of Kalman gain and subsequently, the state and covariance update:   
\begin{align}
K_{k+1}
&=
P^{xy}_{k+1}\left(P^{\nu\nu}_{k+1}\right)^{-1} \\[6pt]
\hat{\boldsymbol{x}}_{k+1}^{+}
&=
\hat{\boldsymbol{x}}_{k+1}^{-}
+
K_{k+1}\left(\boldsymbol{y}_{k+1,\mathrm{meas}}
-
\hat{\boldsymbol{y}}_{k+1}^{-}\right) \\[6pt]
P_{k+1}^{+}
&=
P_{k+1}^{-}
-
P^{xy}_{k+1}K_{k+1}^{\top}
-
K_{k+1}\left(P^{xy}_{k+1}\right)^{\top}
+
K_{k+1}P^{\nu\nu}_{k+1}K_{k+1}^{\top}
\end{align}
With the updated mean error GRP and the \textit{a priori} mean quaternion, the \textit{a posteriori} mean quaternion can be computed as:
\begin{align}
\hat{\boldsymbol{q}}_{k+1}^{+}
&=
\delta\hat{\boldsymbol{q}}_{k+1}^{+}
\otimes
\hat{\boldsymbol{q}}_{k+1}^{-(0)}
\end{align}
Following the multiplicative injection, the attitude error state is reset to zero prior to the next filter cycle:
\begin{align}
\delta\hat{\boldsymbol{p}}_{k+1}^{+} \leftarrow \boldsymbol{0}
\end{align}

\section{Observer Architecture Optimization Framework (Task 1)}\label{sec:arch_opt}
In this section, observer architecture optimization is defined as the joint selection of (i) the orbital trajectories used by the observers and (ii) the number of sensing platforms assigned to each trajectory. Based on prior work in cislunar mission design and space-based surveillance, a set of thirteen periodic orbit families in the Earth-Moon system is considered. These include the northern and southern Butterfly and Dragonfly families, northern and southern Halo orbits about the $L_1$, $L_2$, and $L_3$ libration points, as well as Lyapunov orbits associated with the same three equilibrium points.

\begin{figure}[!ht]
  \begin{center}
  \includegraphics[width=0.4\linewidth]{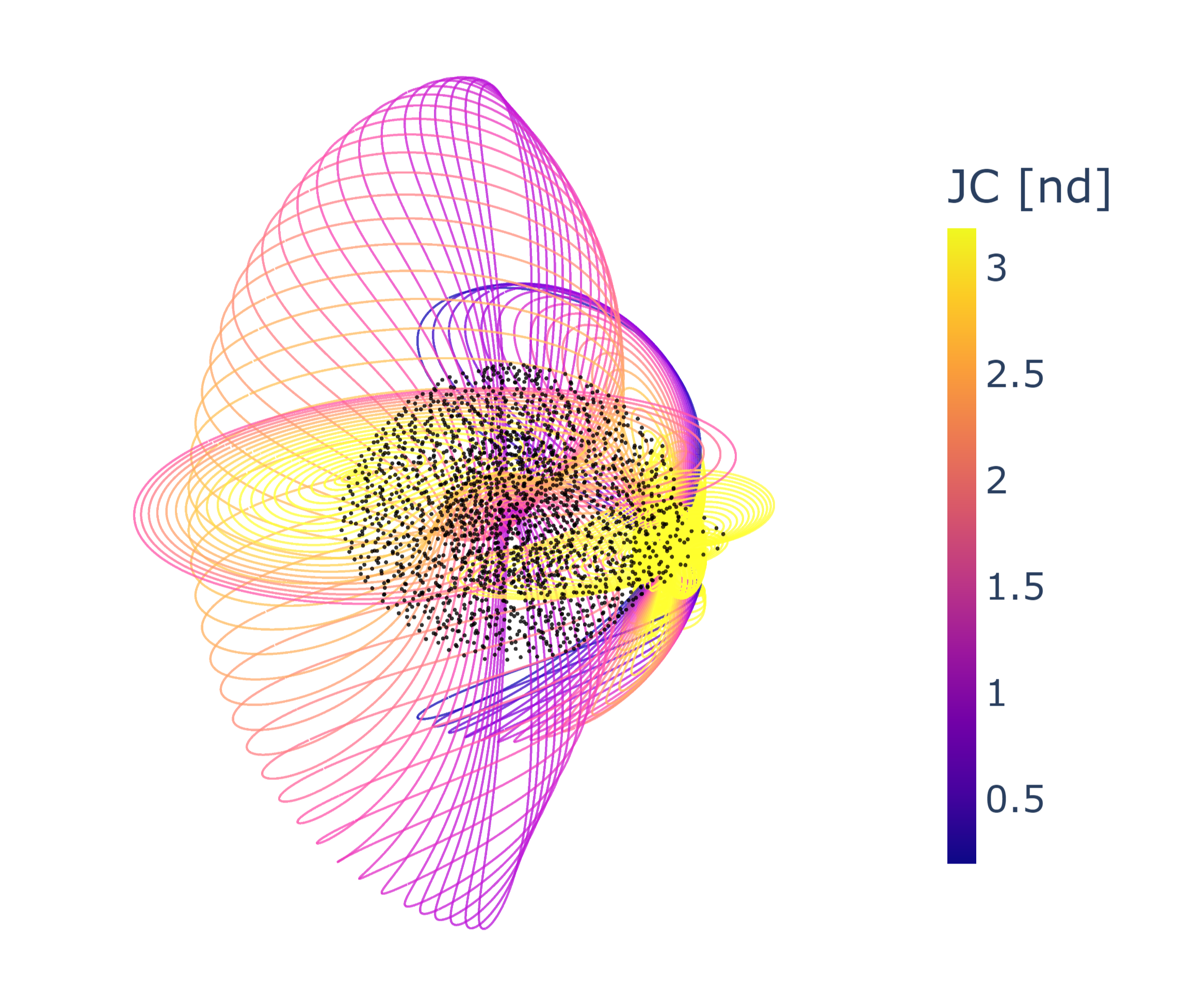}
    \caption{Candidate Observer Orbit Set Consisting of 302 Periodic Trajectories Drawn from 13 Earth-Moon Orbit Families, Shown Together with Fixed Targets in the Rotating Frame (2000 Points).}
  \label{fig:orb_fam}
  \end{center}
\end{figure}

The selected orbit families provide diverse cislunar surveillance geometries due to their differing spatial extents and out-of-plane characteristics. For each family, individual periodic solutions are generated using a continuation procedure in the x-coordinate, initiated at perpendicular crossings of the x-axis. On average, approximately 23 periodic orbits are retained from each family, yielding a total of 302 candidate observer trajectories (Fig.~\ref{fig:orb_fam}). The separation in x among successive orbits varies by family, ranging from approximately 500 km for Dragonfly orbits to nearly 30,000 km for Halo families capable of enveloping both the Earth and Moon. It is worth noting that the step sizes employed during continuation are typically much smaller than these values to ensure numerical convergence.


From the full library of 302 candidate observer orbits, the optimizer simultaneously selects exactly 10 observer trajectories and assigns an integer number of sensing satellites to each selected orbit. The allowable number of observers per orbit ranges from 1 to 10, with satellites distributed along the trajectory using equal temporal offsets. For example, in the case of five observers on a given orbit, the satellites are initialized with relative temporal separations of $T/5$, $2T/5$, $3T/5$, and $4T/5$ with respect to the first satellite, where $T$ denotes the orbital period. The architecture optimization is carried out using a set of hypothesized targets that are fixed in the CR3BP rotating frame. In contrast, the sensor tasking problem (see Section~\ref{sec:sensor_Tk}) considers a different set of targets that move according to CR3BP dynamics.

The set of static target locations used in the architecture optimization is derived from publicly available information on past and planned cislunar missions. A total of 64 missions, covering both government and commercial activities, are identified from public sources \citep{batcha_artemis, williams_nrho, lai_lunar_flashlight}, with associated Jacobi constants spanning from 2.91 (Artemis~I) to 5.49 (several lunar lander missions). For each mission, the Jacobi constant, either computed using $\mathrm{JC}=2U-v^2$ or taken directly from the literature, is used to determine the corresponding zero-velocity surface. Static target points are then sampled in an equiangular manner within each zero-velocity surface. Combining the target sets associated with all Jacobi values results in a collection of fixed points distributed throughout the cislunar region. Figure~\ref{fig:orb_fam} illustrates the fixed points and all 302 candidate observer orbits together.

Regions of the cislunar space associated with Jacobi values shared by multiple missions naturally contain a higher concentration of fixed target points. To obtain a manageable yet representative target set, all sampled points are subsequently clustered using the $k$-means algorithm \citep{LIKAS2003451}, resulting in a user-specified number of targets fixed in the CR3BP frame. The resulting target locations capture both the distribution of anticipated cislunar activity and the associated dynamically accessible regions consistent with the energy levels of space objects, including scenarios involving disposal or anomalous events.

After consolidating the target points using $k$-means clustering, the architecture optimization is evaluated under three target-density cases. The first case, referred to as Scenario~A (``sparse-opt''), uses 100 static targets. Scenario~B (``moderate-opt'') increases the target count to 2000, while Scenario~C (``dense-opt'') further expands the set to 10,000 static targets. In all cases, the targets are fixed in the CR3BP rotating frame. Optimization performance is evaluated using a novel composite objective function introduced next. The proposed cost function, $J$, is partially informed by the framework in \cite{Visonneau2023}. It is defined as the absolute value of the product of six scalar performance terms:
\begin{equation}\label{J_constraint_fn}
    J = \left| \prod_{j=1}^{6} \lambda_j \right|
\end{equation}
where each factor $\lambda_j$ is defined below:
\begin{enumerate}
\item[$\bullet$] $\lambda_1$: this term penalizes architectures that deploy a large number of observer satellites by taking the inverse of the total observer count:
\begin{equation}
\lambda_1 = \left(\sum_{i=1}^{10} n_i\right)^{-1}
\end{equation}
where $n_i$ denotes the number of observer satellites assigned to orbit $i$. 
\item[$\bullet$] $\lambda_2$: this factor accounts for orbital stability by weighting the number of observer satellites on each orbit with a stability index:
\begin{equation}
\lambda_2 = \left(\sum_{i=1}^{10} n_i\,\Xi_i\right)^{-1}
\end{equation}
where $\Xi_i$ denotes the stability index of orbit $i$, defined from the eigenvalue $\eta_i$ of the corresponding monodromy matrix as $\Xi_i = \tfrac{1}{2}|\eta_i| + \tfrac{1}{2}|\eta_i|^{-1}$. Values of $\Xi_i$ greater than unity indicate linear instability.
\item[$\bullet$] $\lambda_3$: this term quantifies cumulative target observability over a fixed propagation horizon. Observer trajectories are propagated for $t_{\textrm{propagation}}=30$ days, with observations evaluated at 1-hour intervals starting from the initial epoch $t=0$. 
\begin{equation}
\lambda_3 =
\frac{1}{n_{\textrm{targets},s}\,n_{\textrm{steps}}}
\sum_{t_k}\sum_{i=1}^{10}\sum_{j=1}^{n_i}
V_{t_k,i,j}
\end{equation}
where $V_{t_k,i,j}$ denotes the number of targets observable from satellite $j$ on orbit $i$ at time $t_k$ ($t_k = 0, 1\mathrm{H}, 2\mathrm{H}, \ldots, t_{\textrm{propagation}}$). A target is considered observable only if its apparent brightness (Eq.~\ref{eq:mag_sph}) exceeds a prescribed threshold and Sun, Moon, and Earth exclusion-angle constraints of $35^{\circ}$, $5^{\circ}$, and $15^{\circ}$, respectively, are satisfied. The quantities $n_{\textrm{targets},s}$ and $n_{\textrm{steps}}$ denote the number of static targets and time steps over the propagation horizon, respectively, and are included solely to normalize the magnitude of $\lambda_3$. 
\item[$\bullet$] $\lambda_4$: While $\lambda_3$ characterizes the total number of observable targets, it offers no insight into the number of times each target remains observable. $\lambda_4$ represents the number of times each target remains observable, scaled to remove bias from the number of observer satellites:
    \begin{align}
\lambda_4&=\frac{1}{n_{\textrm{targets},s}n_{steps}}\sum_{l=1}^{n_{\textrm{targets},s}}\log_{10}\left(\sum_{i=1}^{10}\frac{\sum_{t_k}\sum_{j=1}^{n_i}\mathcal{V}_{t_k,i,j,l}}{n_i}\right)
    \end{align}
where $\mathcal{V}_{t_k,i,j,l} \in \{0,1\}$ denotes an observability indicator, taking the value 1 when target $l$ is observable to satellite $j$ of orbit $i$ at time $t_k$, and 0 otherwise. For observer orbit $i$, the total number of times target $l$ is observable over the full time horizon is normalized by the satellite count $n_i$ associated with that orbit. The $\log_{10}$ transformation is introduced to prevent unbounded growth and to exploit the logarithmic identity $\log_{10}\!\left(\prod_k a_k\right)=\sum_k \log_{10}(a_k)$, which converts multiplicative factors into additive terms. The same brightness-threshold and exclusion-angle constraints described previously are applied to determine observability. The normalization factor $n_{\textrm{targets},s}n_{\textrm{steps}}$ serves only to scale the metric. To avoid undefined values, any zero argument of the logarithm is replaced with unity.
\item[$\bullet$] $\lambda_5$: A proximity metric is introduced to penalize large distances from both the Earth and the Moon, thereby favoring observer configurations that remain close to the two primary bodies. The metric is based on the cumulative Earth and Moon-relative distances of all observer satellites over the propagation horizon:
\begin{align}
\lambda_5^{-1}
&= \frac{1}{n_{\textrm{targets},s} n_{\textrm{steps}}}
\sum_{t_k}\sum_{i=1}^{10}\sum_{j=1}^{n_i}
D_{t_k,i,j}^{\textrm{tot}} \\
D_{t_k,i,j}^{\textrm{tot}}
&\triangleq D_{t_k,i,j,\textrm{Moon}} + D_{t_k,i,j,\textrm{Earth}} 
\end{align}
where, $D_{t_k,i,j,\textrm{Moon}}$ and $D_{t_k,i,j,\textrm{Earth}}$ denote the distances from satellite $j$ in orbit $i$ to the Moon and Earth, respectively, at time $t_k$. The normalization factor $n_{\textrm{targets},s}n_{\textrm{steps}}$ serves only to scale the metric magnitude.
\item[$\bullet$] $\lambda_6$: It is a binary feasibility term incorporated into the cost function to enforce orbit uniqueness. The optimization framework is constrained to select ten distinct observer orbits. The metric evaluates to $1$ when all selected orbits are unique and to $0$ if any duplicate orbits are present.
\end{enumerate}

To minimize the composite cost function $J$ over the resulting design space, this work uses the open-source Python package Hyperopt \citep{hyperopt}. The package is well suited for optimization problems with many coupled decision variables and provides a systematic and reproducible alternative to manual parameter tuning. Hyperopt includes three search algorithms: Random Search, the Tree-structured Parzen Estimator (TPE), and the Adaptive TPE (ATPE), with the latter two belonging to the Bayesian optimization family. The three search strategies are briefly outlined below:
\begin{enumerate}
    \item[$\bullet$] Random Search: This method \citep{bergstra2012random} serves as a baseline by sampling candidate configurations uniformly from the predefined discrete design space. Each sampled configuration is evaluated by computing the objective cost $J$. After a fixed number of trials, the configuration with the best observed cost is selected.
\item[$\bullet$] Tree-Structured Parzen Estimator: TPE \citep{bergstra2011algorithms} is a sequential model-based approach that constructs probabilistic models from prior evaluations. Evaluated configurations are partitioned according to their objective costs, with the lower-cost group modeled by the density $l(x)$ and the remaining configurations by $g(x)$. New candidates are then selected by favoring regions where the ratio $l(x)/g(x)$ is large, directing the search toward lower-cost configurations. The density models are updated after each evaluation, progressively concentrating the search and typically requiring fewer cost-function evaluations than random sampling.
\item[$\bullet$] Adaptive TPE: ATPE extends TPE by adaptively adjusting how past evaluations are partitioned when estimating the densities $l(x)$ and $g(x)$, rather than using fixed quantile thresholds. This adaptive modeling can improve search efficiency. \end{enumerate}

In this work, TPE is used as the primary optimizer, with Random Search serving as a baseline for comparison.

\section{Integrated Sensor Tasking and State Estimation (Task 2)}\label{sec:sensor_Tk}
Sensor tasking uses the observer architectures obtained from the preceding optimization, with the selected orbits and satellite counts fixed for each of Scenarios~A, B, and C. The target set is drawn from five cislunar orbit families and trajectories associated with several lunar missions: northern and southern Halo orbits about the $L_1$ and $L_2$ libration regions (with two near-rectilinear Halo orbits), distant retrograde orbits, 3:1 resonant orbits, and eight additional trajectories with perigee altitudes of a few hundred kilometers above the lunar south pole~\citep{chandrayaan_1_a,chandrayaan_2_a,chandrayaan_3_a,kplo_a,kerner_lunah,genova_lunah_traj}. In total, 202 targets distributed across these orbit classes are considered.

For computational tractability, the analytic brightness model (Section~\ref{chap:analytical_brightness}) is used in place of high-fidelity facet-based photometric simulations. Each target is modeled as a uniformly reflective sphere, allowing closed-form magnitude calculations. The sensor tasking problem is formulated as a single-step (i.e, greedy) optimization and re-solved at discrete decision epochs using a fixed update cadence (e.g., every two hours). Let $N_o$ denote the number of observers and $N_t$ be the number of candidate targets. Enforcing a one-to-one assignment constraint, the number of feasible configurations at each epoch is ${}^{N_t}P_{N_o}$. For example, with $N_o = 26$ sensors and $N_t = 175$ potential objects, this results in ${}^{175}P_{26}$ distinct allocation possibilities. The resulting search space is too large for exhaustive evaluation, so the tasking optimization is carried out using a Bayesian approach based on Hyperopt’s TPE algorithm. Assuming independence between the orbital state estimates of the assigned targets, we define the joint covariance at time $t_k$ as:
\begin{equation}
\hat{P}_{\mathrm{joint},t_k}
=
\operatorname{blkdiag}\!\left[
\hat{P}_{\ell_{1 j_1}, t_k}\;,\;
\hat{P}_{\ell_{2 j_2}, t_k}\;,\;
\dots\;,\;
\hat{P}_{\ell_{n_T j_{n_T}}, t_k}
\right]
\end{equation}
where $\hat{P}_{\ell_{i j_i}, t_k} \in \mathbb{R}^{6\times 6}$ denotes the orbital state covariance of target $i$ assigned to sensor $j_i$ at time $t_k$. The joint covariance is block diagonal, with zero off-diagonal blocks. At each decision epoch, sensor assignments are selected by maximizing the mutual information metric \citep{doi:10.2514/6.2024-1676}:
\begin{equation}\label{MI_joint}
\mathcal{M}_{\mathrm{joint},t_k}
=
\frac{1}{2}
\log\!\left(
\frac{\det\!\left(\hat{P}_{\mathrm{joint},t_k}^{-}\right)}
{\det\!\left(\hat{P}_{\mathrm{joint},t_k}^{+}\right)}
\right)
\end{equation}
where, $\hat{P}_{\mathrm{joint},t_k}^{-}$ and $\hat{P}_{\mathrm{joint},t_k}^{+}$ are assembled from the \textit{a priori} and \textit{a posteriori} target covariances $\hat{P}_{\ell_{i j_i},t_k}^{-}$ and $\hat{P}_{\ell_{i j_i},t_k}^{+}$, respectively. Because the matrix is block diagonal, its determinant decomposes as $\det(\hat{P}_{\mathrm{joint},t_k}) = \prod_{i=1}^{n_T} \det(\hat{P}_{\ell_{i j_i},t_k})$, allowing the mutual information metric to be computed directly from the individual target covariances without forming the full joint matrix. To obtain the \textit{a posteriori} covariance matrix, we use the Joseph form:
\begin{equation}
\hat{P}_{ij_it_k}^{+}
=
\hat{P}_{ij_it_k}^{-}
-
K_{k_{j_i}} C_{k_{j_i}}^{T}
-
\left(K_{k_{j_i}} C_{k_{j_i}}^{T}\right)^{T}
+
K_{k_{j_i}} W_{k_{j_i}} K_{k_{j_i}}^{T}
\end{equation}
where $W_{k_{j_i}}$ denotes the predicted covariance in the measurement space,
$C_{k_{j_i}}$ is the state-measurement cross-covariance, and the Kalman gain is
$K_{k_{j_i}} = C_{k_{j_i}} W_{k_{j_i}}^{-1}$.

Conditioned on the sensor-target assignments determined by the sensor tasking optimization, state estimation proceeds at a substantially finer cadence (30 seconds in this work), whereas the tasking assignments are updated at coarse intervals. The estimated state vector includes translational and rotational components. The associated measurements consist of line-of-sight pointing directions together with photometric brightness observations. As a simplification, orbital and attitude dynamics are assumed decoupled. The estimation problem is addressed using an error-state multiplicative UKF formulation \citep{doi:10.2514/2.5102}, which employs GRPs for attitude-error representation.

\section{Results and Discussions}\label{sec:results}
\subsection{Architecture Optimization (Task 1) Results}\label{sec:AOresults}
The numerical simulation parameters for the architecture optimization are provided in Table \ref{arch_opt_settings}. For an observer orbit $i$ hosting $n_i$ number of satellites, the position of the first satellite at the initial epoch is specified by $y_{CR3BP}=0$. Since the observer orbits are propagated over a 30-day interval, and many have substantially shorter orbital periods, numerical error accumulation can result in orbital divergence. In order to address the divergence issue, the observer orbits are wrapped around by their respective orbital periods.  
\begin{table}[htbp]
 \caption{Simulation Settings for the Architecture Optimization}
 \label{arch_opt_settings}
 \centering
 \begin{tabular}{p{0.40\linewidth} | p{0.53\linewidth}}
 \hline
 \hline
  Parameter & Value or Description\\
 \hline
Initial epoch & 2460584.5 JD (00:00:00 UT, October 1, 2024)\\
Orbit propagation method & Runge-Kutta order 5(4) \citep{DORMAND198019}\\
Integration tolerances & Relative and absolute tolerances $=10^{-10}$\\
Static target shape and dimension & Sphere with 1 $m$ radius \\
Maximum number of architecture evaluations & 10,000\\
Optimization early stopping criterion & 500 iterations\\
Visibility threshold & 18 $mag$\\
 \hline
 \hline
 \end{tabular}
 \end{table}
 
\begin{figure}[h]
     \centering
     \begin{subfigure}[h]{0.49\linewidth}
         \centering
         \includegraphics[width=\linewidth]{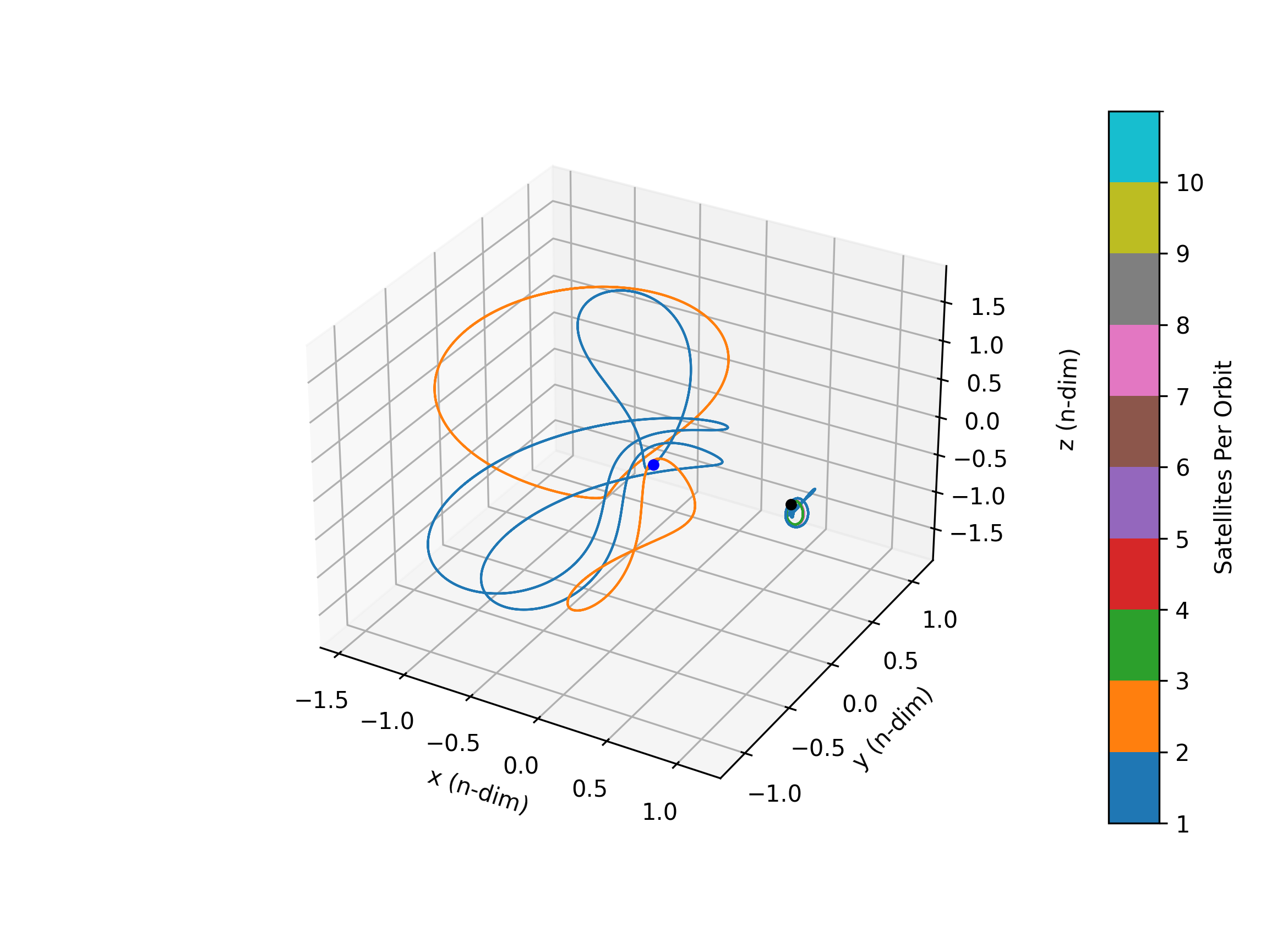}
         \caption{Observer Orbits for 100 Static Targets}
         \label{sel_orbits_tpe_1}
     \end{subfigure}
     \begin{subfigure}[h]{0.49\linewidth}
         \centering
         \includegraphics[width=\linewidth]{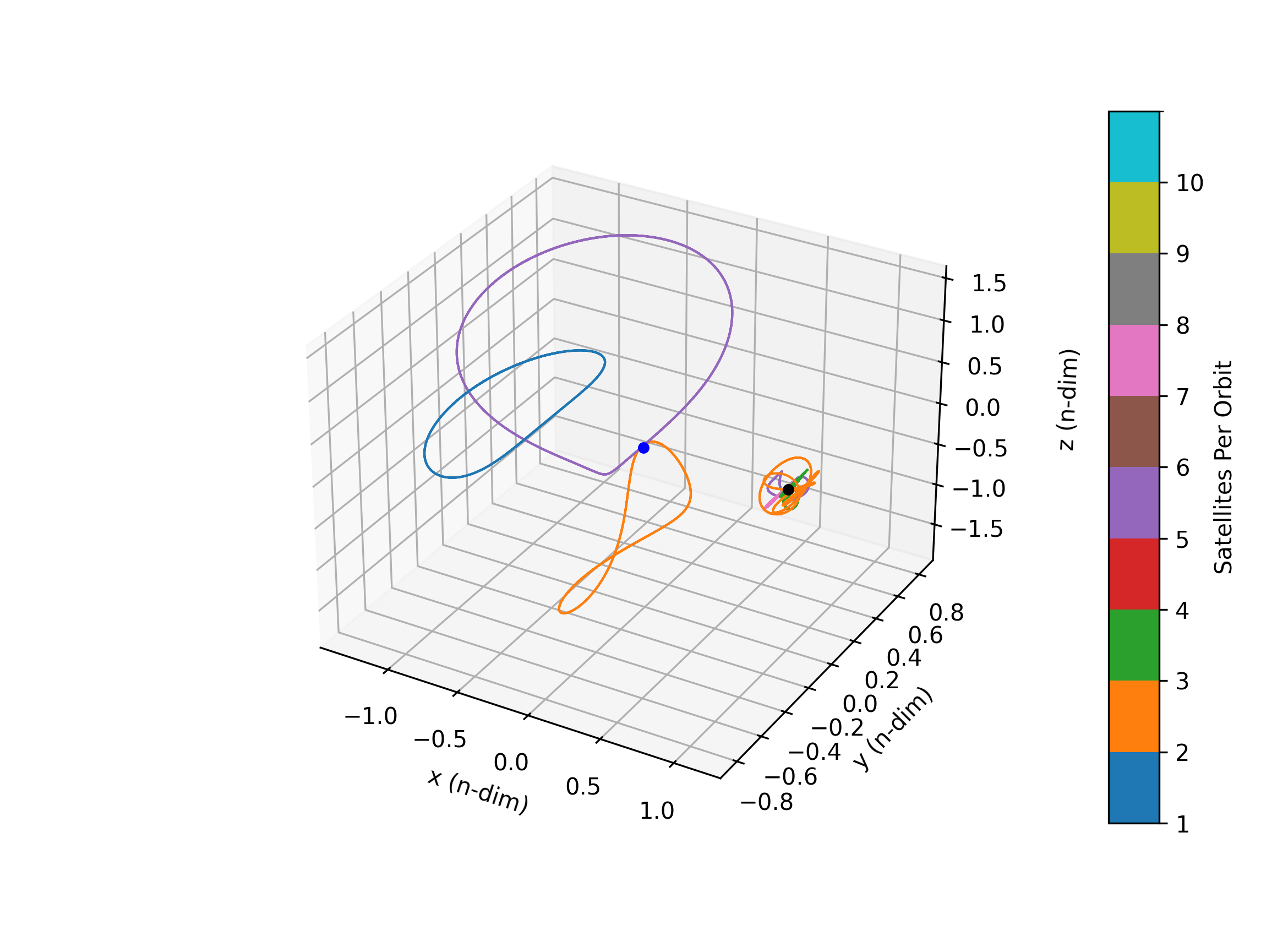}
         \caption{Observer Orbits for 2000 Static Targets}
         \label{sel_orbits_tpe_2}
     \end{subfigure}   
     \begin{subfigure}[h]{0.49\linewidth}
         \centering
         \includegraphics[width=\linewidth]{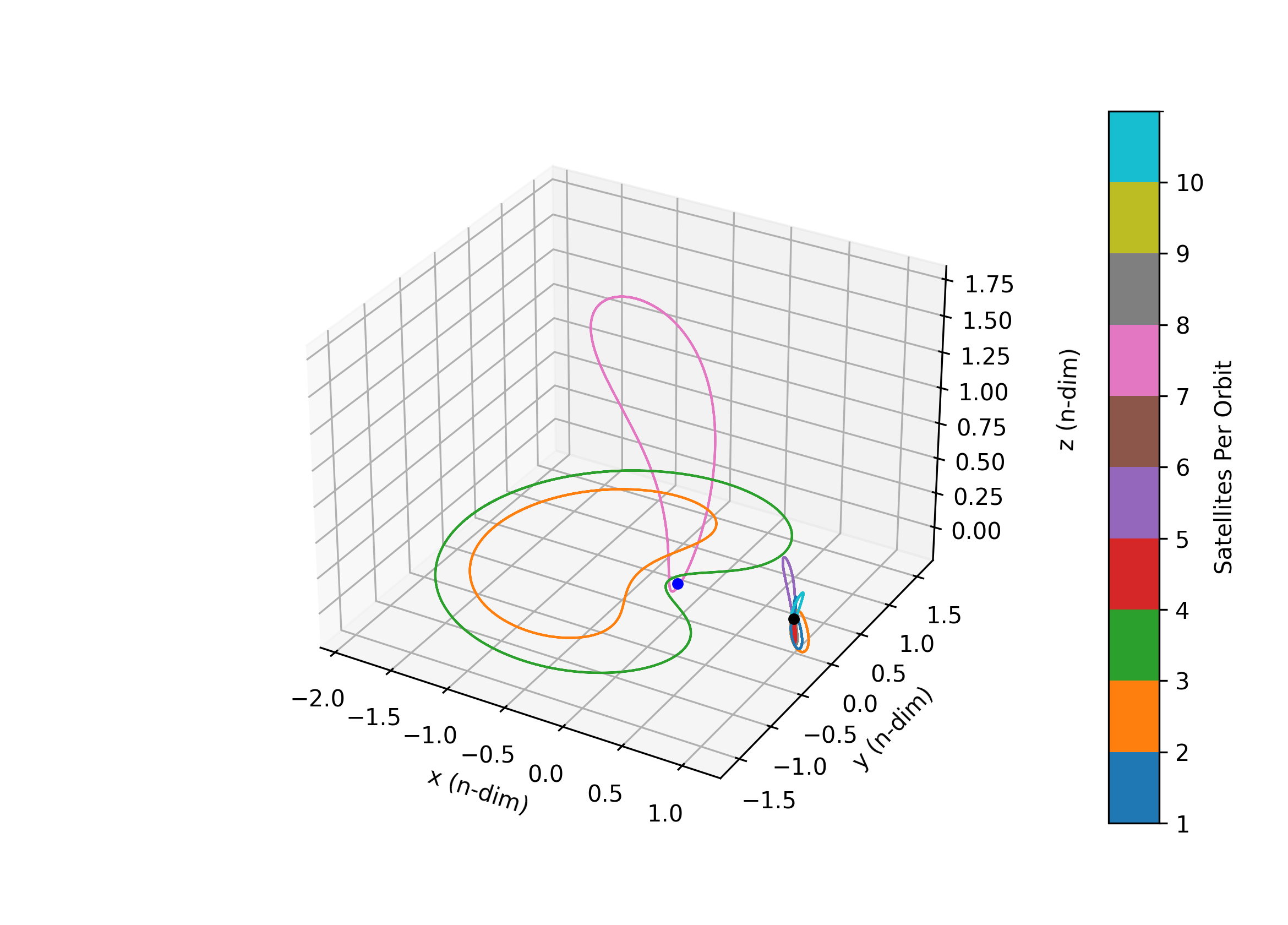}
         \caption{Observer Orbits for 10000 Static Targets}
         \label{sel_orbits_tpe_3}
     \end{subfigure}        
        \caption{Observer Orbits Selected by the TPE-Based Optimizer for (a) Sparse-Opt (Scenario A), (b) Moderate-Opt (Scenario B), and (c) Dense-Opt (Scenario C).}

        \label{sel_orbits_tpe}
\end{figure}

\begin{figure}[h]
     \centering
     \begin{subfigure}[h]{0.49\linewidth}
         \centering
         \includegraphics[width=\linewidth]{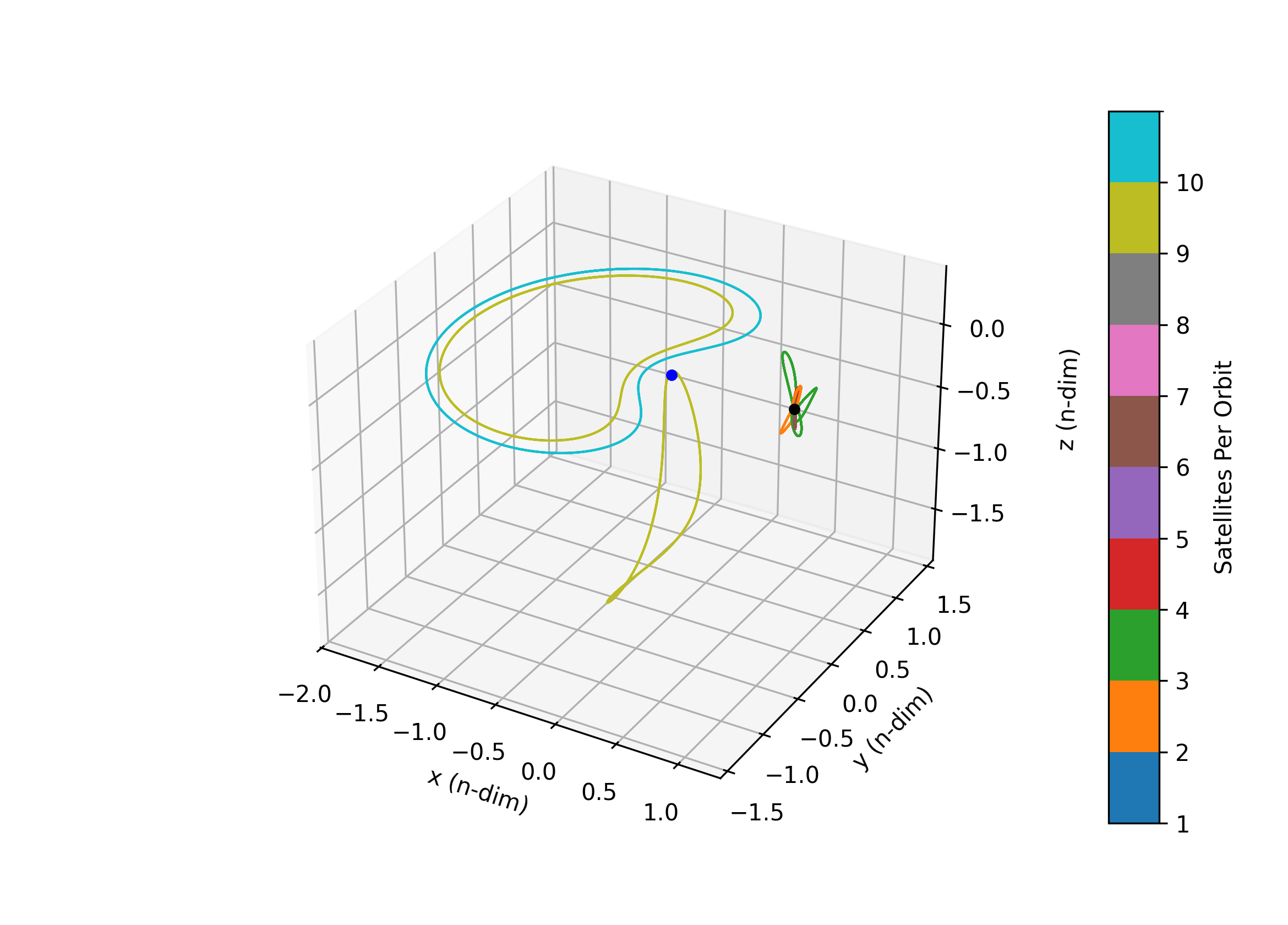}
         \caption{Observer Orbits for 100 Static Targets}
         \label{sel_orbits_rand_1}
     \end{subfigure}
     \begin{subfigure}[h]{0.49\linewidth}
         \centering
         \includegraphics[width=\linewidth]{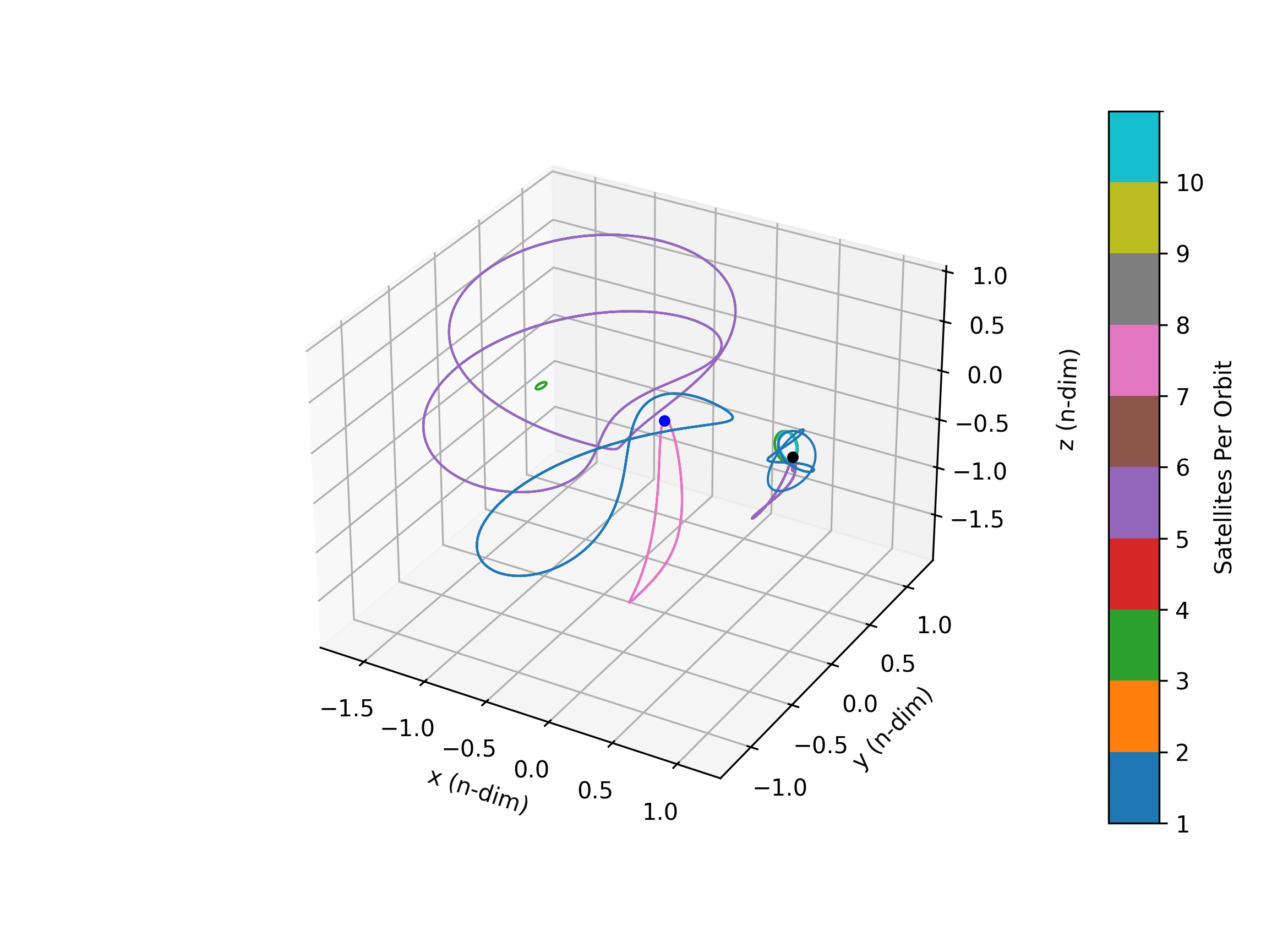}
         \caption{Observer Orbits for 2000 Static Targets}
         \label{sel_orbits_rand_2}
     \end{subfigure}   
     \begin{subfigure}[h]{0.49\linewidth}
         \centering
         \includegraphics[width=\linewidth]{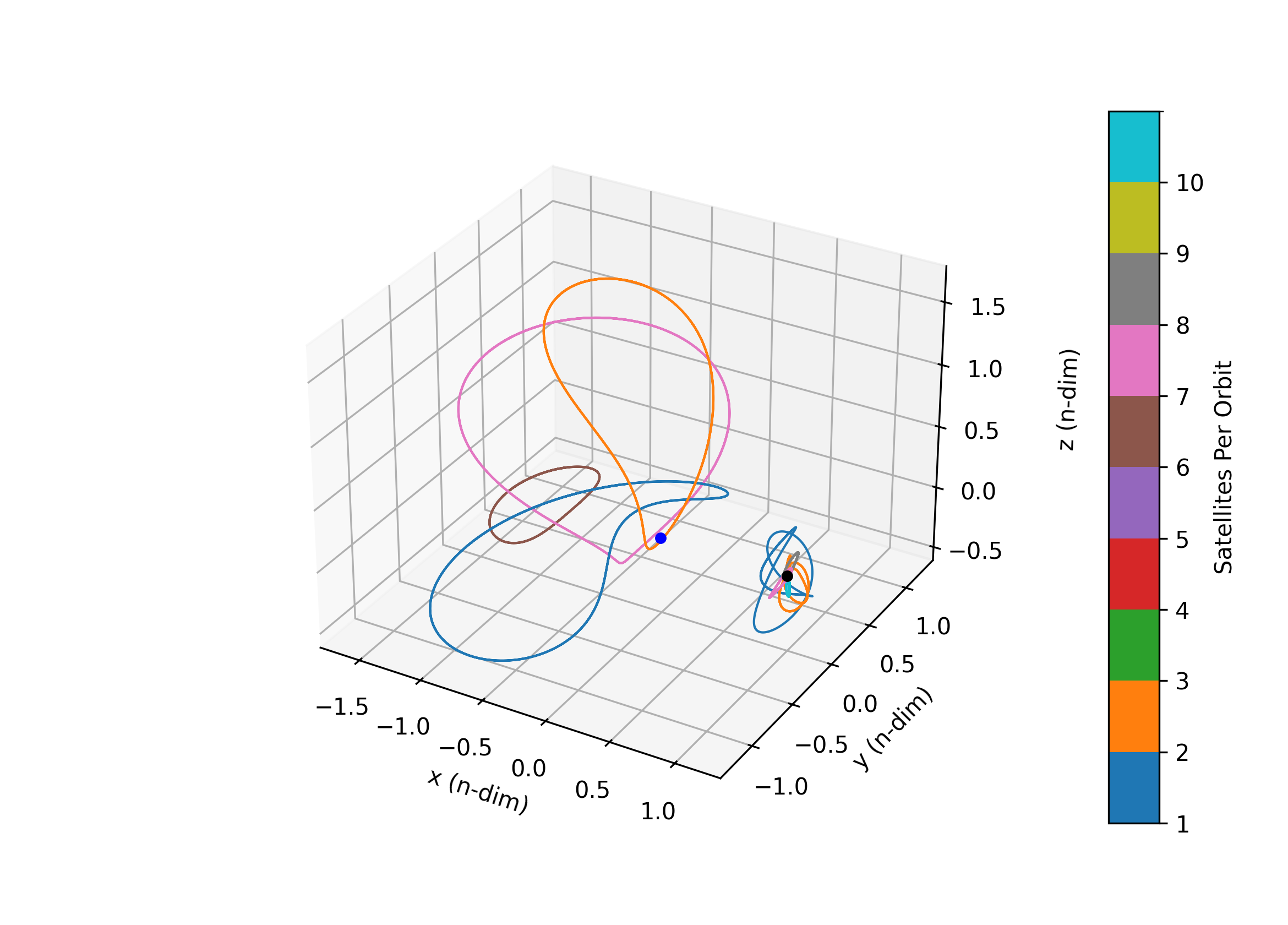}
         \caption{Observer Orbits for 10000 Static Targets}
         \label{sel_orbits_rand_3}
     \end{subfigure}        
        \caption{Observer Orbits Selected by the Random Search-Based Optimizer for (a) Sparse-Opt (Scenario A), (b) Moderate-Opt (Scenario B), and (c) Dense-Opt (Scenario C).}        
        \label{sel_orbits_rand}
\end{figure}

Table \ref{arch_opt_allcases} presents the orbit families and corresponding cardinalities (number of satellites) for the ten observer orbits associated with the best architectures selected by the TPE-based optimizer. Figure~\ref{sel_orbits_tpe} illustrates the orbits, shown in the non-dimensional CR3BP frame, corresponding to the best architectures obtained from the TPE-based optimization process. Figures~\ref{sel_orbits_tpe_1}-\ref{sel_orbits_tpe_3} present the results for Scenarios A, B, and C, respectively, with the Earth and Moon denoted by blue and black markers. Under the TPE-based optimization framework, the best architecture comprises 20 observer satellites for Scenario A, 33 observers for Scenario B, and 43 observers for Scenario C. For comparison, Fig.~\ref{sel_orbits_rand} shows the orbits of the best architecture selected by the random search-based optimizer, highlighting clear differences relative to the TPE-based results. The best architectures for Scenarios A, B, and C identified by Random Search comprise 51, 45, and 46 satellites, respectively. Furthermore, as expected, TPE consistently outperforms Random Search across all scenarios. For Scenario A, TPE achieves a cost function value of 1.5081$\times 10^{-4}$ compared to Random Search's 1.2236$\times 10^{-5}$. Similarly, in Scenario B, TPE yields 3.9069$\times 10^{-4}$ versus Random Search's 1.1624$\times 10^{-4}$. For Scenario C, TPE attains 3.4424$\times 10^{-3}$ while Random Search achieves only 6.0012 $\times 10^{-4}$. Notably, Random Search achieves lower cost function values despite utilizing more satellites than TPE, demonstrating TPE's superior exploration capabilities.
\begin{table}[htbp]
\caption{Best-Performing Architectures Identified by the TPE-Based Optimizer for the Sparse-Opt (Scenario A), Moderate-Opt  (Scenario B), and Dense-Opt (Scenario C) Cases}
\label{arch_opt_allcases}
 \centering
 \small
\setlength{\tabcolsep}{3pt}
 \begin{tabular}{c | c | c}
 \hline
 \hline
 Orbit Number & \makecell{Orbit Family\\ Scenario A | B | C } &\makecell{ Number of Satellites\\ Scenario A | B | C } \\
 \hline
\begin{tabular}{c}
1 \\ 2 \\ 3 \\ 4 \\ 5 \\6\\7\\8\\9\\10
\end{tabular}
&
\begin{tabular}{@{}l c l c l@{}}
$L_3$ Halo north & \textbar & Butterfly north & \textbar & $L_1$ Halo north   \\
$L_3$ Halo south & \textbar & $L_3$ Halo south & \textbar & $L_2$ Halo south   \\
$L_2$ Halo south & \textbar & $L_1$ Halo south & \textbar & $L_3$ Halo north   \\
$L_3$ Halo north & \textbar & $L_3$ Halo north & \textbar & $L_2$ Halo north   \\
$L_3$ Halo south & \textbar & $L_2$ Halo south & \textbar & $L_2$ Halo north   \\
$L_2$ Halo south & \textbar & $L_2$ Halo south & \textbar & $L_2$ Halo south   \\
$L_3$ Halo south & \textbar & $L_2$ Halo north & \textbar & $L_3$ Lyapunov   \\
$L_2$ Halo north & \textbar & Butterfly south & \textbar & $L_2$ Halo south  \\
$L_2$ Halo south & \textbar &  Butterfly south & \textbar & $L_2$ Halo south   \\
$L_2$ Halo south & \textbar & $L_3$ Lyapunov & \textbar & $L_3$ Lyapunov   \\
\end{tabular}
&
{\setlength{\tabcolsep}{3pt}
\begin{tabular}{@{}l c l c c@{}}
2 & \textbar & 5 & \textbar & 5 \\
1 & \textbar & 2 & \textbar & 2 \\
7 & \textbar & 7 & \textbar & 7 \\
1 & \textbar & 5 & \textbar & 1 \\
1 & \textbar & 3 & \textbar & 10 \\
1 & \textbar & 3 & \textbar & 8 \\
2 & \textbar & 3 & \textbar & 2 \\
1 & \textbar & 2 & \textbar & 4 \\
3 & \textbar & 2 & \textbar & 1 \\
1 & \textbar & 1 & \textbar & 3 \\
\end{tabular}}
\\
\hline
\hline
\end{tabular}
\end{table}


\subsection{Sensor Tasking and Estimation (Task 2) Results}\label{sec:STEresults}
Utilizing the best architectural outcomes obtained from Scenarios A, B, and C in Task 1, we investigate a combined sensor tasking-state estimation problem. Table \ref{st_se_settings} lists the key simulation parameters. For each target, the initial mean translational state (position and velocity) is generated by propagating the target from $y_{CR3BP}=0$ for a randomly selected fraction of its orbital period. The initial translational truth state is then simulated by sampling a state vector from the initial translational distribution defined by the target’s mean and covariance (see Table~\ref{st_se_settings}). This initial translational truth state is propagated using CR3BP dynamics to obtain the translational truth states over the entire simulation horizon. 

The initial truth quaternion is simulated through the following process: (1) an attitude error vector is sampled from a zero-mean multivariate Gaussian distribution with covariance defined by the attitude error variance (see Table~\ref{st_se_settings}), (2) this attitude error is converted to an error quaternion, and (3) a quaternion product between the error quaternion and the initial mean quaternion estimate is performed to generate the initial truth quaternion. The initial truth angular velocity is obtained by drawing a sample from a Gaussian distribution centered at the target’s initial mean angular velocity, with covariance defined by the angular velocity variance (see Table~\ref{st_se_settings}). The initial truth quaternion and angular velocity are propagated using quaternion kinematics (Eq. 2.38, \cite{guzzetti2016coupled}) and torque-free Euler's equations of motion (Eq. 2.39, \cite{guzzetti2016coupled}) to obtain the rotational truth states over the entire simulation period.

\begin{table}[htbp]
 \caption{Simulation Settings for the Dual Sensor Tasking-State Estimation Task}
 \label{st_se_settings}
 \centering
 \begin{tabular}{p{0.35\linewidth} | p{0.6\linewidth} }
 \hline
 \hline
  Parameter & Value or Description\\
 \hline
Initial epoch & 2460584.5 JD (00:00:00 UT, October 1, 2024)\\
Orbit propagation method & Runge-Kutta 5(4) with relative/absolute tolerances: $10^{-10}$\\
Orbit propagation period & 1 day\\
Targets' initial covariance & Diagonal with position variance $10^{-1}\times 100000^2$ $m^2$\\
& Velocity variance $10^{-1}\times (0.1)^2$ $(m/s)^2$\\
& attitude error variance $5^2$ $deg^2$\\ 
& angular velocity variance $(10^{-1}\times 24)^2$ $(deg/hr)^2$ \\
Targets' initial mean quaternion & $[0,0,0,1]^T$\\
Targets' initial mean angular velocity & $[-44.723808, -6.6573, -8.514216]^T$ $deg/hr$\\
Targets' inertia matrix & $Diagonal(1047.2,1047.2,1047.2)$ $kg$ $m^2$\\
Observers' initial quaternion & $[0,0,0,1]^T$\\
Observers' initial angular velocity & $[0,0,1]^T$ $deg/hr$\\
Observers' inertia matrix & $Diagonal(4000,4000,4000)$ $kg$ $m^2$\\
Measurement/Estimation frequency & Every 60 seconds\\
Sensor tasking frequency & 30 $mins$, 1 $hr$, 2 $hr$, 3 $hr$, 4 $hr$\\
Measurement noise matrix & $Diagonal(0.1^2,{3^{''}}^2,{3^{''}}^2)$\\
Camera field of view (FoV) & $3$ $deg$\\
GRP parameter $a$ & 0.5\\
Observation constraints & Visibility threshold 18 $mag$\\
&Sun/Moon/Earth exclusion angles $35^{\circ}$, $5^{\circ}$, $15^{\circ}$\\
Maximum number of sensor tasking evaluations & 5000 (at each tasking epoch)\\
Sensor tasking optimization early stopping criterion & 700 iterations\\
\hline
 \hline
 \end{tabular}
 \end{table}

Beginning at the initial epoch, sensor tasking is performed at regular intervals (e.g., every 1 hour). Target selection is parameterized by a continuous weight assigned to each candidate target; sorting these weights and selecting the top $K=n_{obs}$ entries yields a set of $K$ distinct targets, ensuring that each observer is assigned a unique target. At each tasking epoch, Hyperopt's TPE maximizes the mutual information (Eq.~\ref{MI_joint}) to determine the near-optimal sensor-target associations. 

Several implementation-specific considerations for the optimization are described next. To mitigate the computational burden of the sensor tasking optimization, the target objects are substituted with representative spheres of equivalent surface area, thereby allowing the analytical brightness model to be used to evaluate estimated target visibility within the telescope FoV. In evaluating the mutual information metric (Eq.~\ref{MI_joint}), the covariance matrix portion corresponding to the angular states are excluded, and only position and velocity uncertainties are used, providing additional computational advantage. The product of the determinants of the selected target covariances may become numerically ill-conditioned, i.e., vanishingly small when non-dimensional or excessively large when dimensional. Therefore, to ensure numerical robustness, the logarithmic identity $\log_{10}(abcd\hdots)=\log_{10}(a) + \log_{10}(b) + \log_{10}(c) + \hdots$ is employed.

After the optimizer determines the `best' target-observer associations, these assignments are held fixed over the sensor tasking interval (e.g., every 1 hour), while filtering updates are performed at a higher rate using measurements of brightness, right ascension, and declination. Although an analytic brightness model/representative sphere is used during sensor tasking for computational efficiency, the filtering stage instead uses the true target shapes (icosahedron, in this study) with a higher-fidelity facet-based Cook-Torrance model for both brightness measurement generation and modeling.

Simulation of the measurements is discussed next. Brightness measurements are simulated by computing a nominal value from the true target position and quaternion and sampling a noisy observation using the variance in Table \ref{st_se_settings}. Likewise, right ascension and declination measurements are simulated by computing the observer-to-true target direction in the observer body frame and sampling noisy angular observations using the measurement variances listed in Table \ref{st_se_settings}. In the implemented framework, a sensor-tasked target produces a valid observation only when (1) its brightness measurement is above the visibility threshold; (2) the right ascension/declination measurements remain within $\pm0.5$ FoV of the expected angles derived from the observer-to-estimated-target direction in the observer body frame; and (3) all Sun, Earth, and Moon exclusion-angle requirements are met. It is worth emphasizing that all computations are vectorized to manage computational load given the large number of observers and targets. Sensor tasking-estimation performance is next analyzed under variations in target count and tasking interval. 

\subsubsection{Sensor Tasking-Estimation Performance: Varying the Number of Targets of Interest}
In this investigation, we assess how well we can estimate the orbital and rotational states as the number of monitored targets increases. The number of monitored targets $n_T$ is roughly selected from a candidate set $\tau_{n_T}= \{20,25,30,50,75,100,125,150,175,$ $202\}$ and is constrained to be no smaller than the number of observers in each scenario: (1) for Scenario A with 20 observers, $\tau_{n_{T_A}}=\{20,25,30,50,75,100,125,150,175,202\}$, (2) for Scenario B with 33 observers, $\tau_{n_{T_B}}=\{33,50,75,100,125,150,175,202\}$, and (3) for Scenario C with 43 observers, $\tau_{n_{T_C}}=\{43,50,75,100,125,150,175,202\}$. To ensure repeatability, a fixed random seed is used to generate the monitored target sets. Furthermore, target sets of larger $n_T$ are constructed as supersets of those of smaller $n_T$. Sensor tasking optimization is performed every 1 hour, i.e., $T_{ST}=1H$.
\begin{figure}[htbp]
     \centering
     \begin{subfigure}[h]{0.49\linewidth}
         \centering
         \includegraphics[width=.7\linewidth]{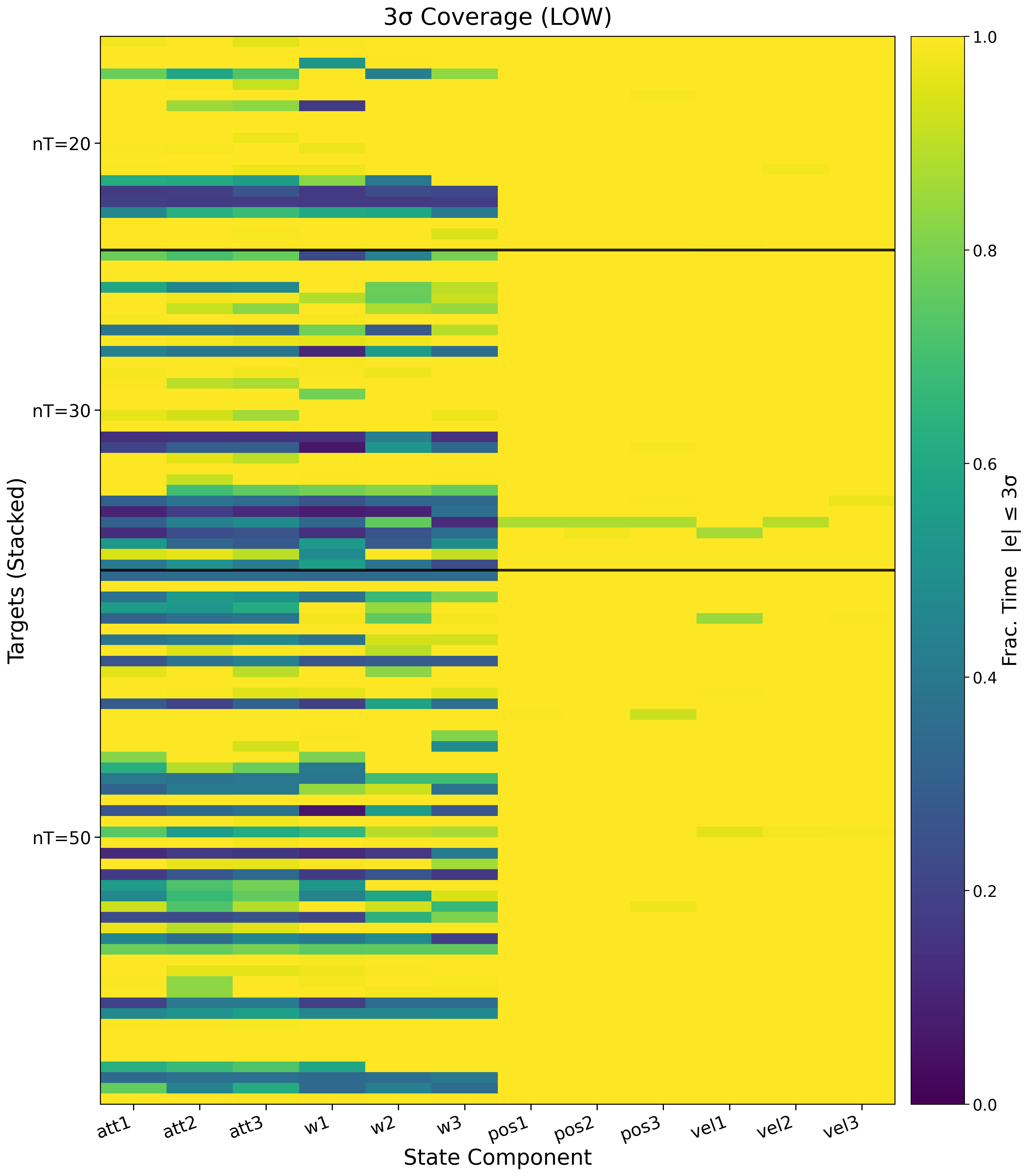}
         \caption{Fraction of Time $|e|\leq 3\sigma$ ($n_T=20,30,50$)}
         \label{allowedtargetstest_scA_1}
     \end{subfigure}
     \begin{subfigure}[h]{0.49\linewidth}
         \centering
         \includegraphics[width=.7\linewidth]{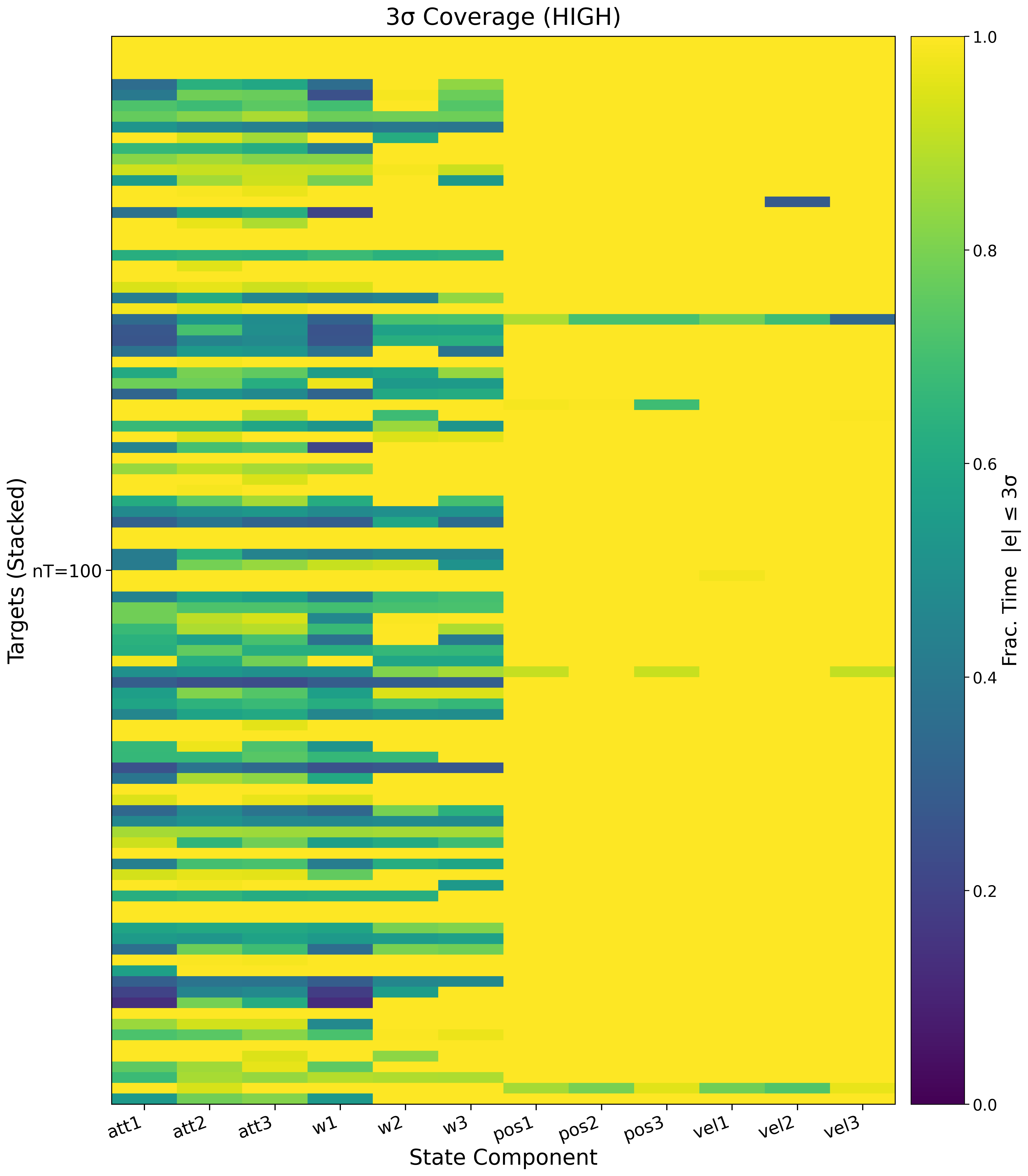}
         \caption{Fraction of Time $|e|\leq 3\sigma$ ($n_T=100$)}
         \label{allowedtargetstest_scA_2}
     \end{subfigure}   
     \begin{subfigure}[h]{0.49\linewidth}
         \centering
         \includegraphics[width=.7\linewidth]{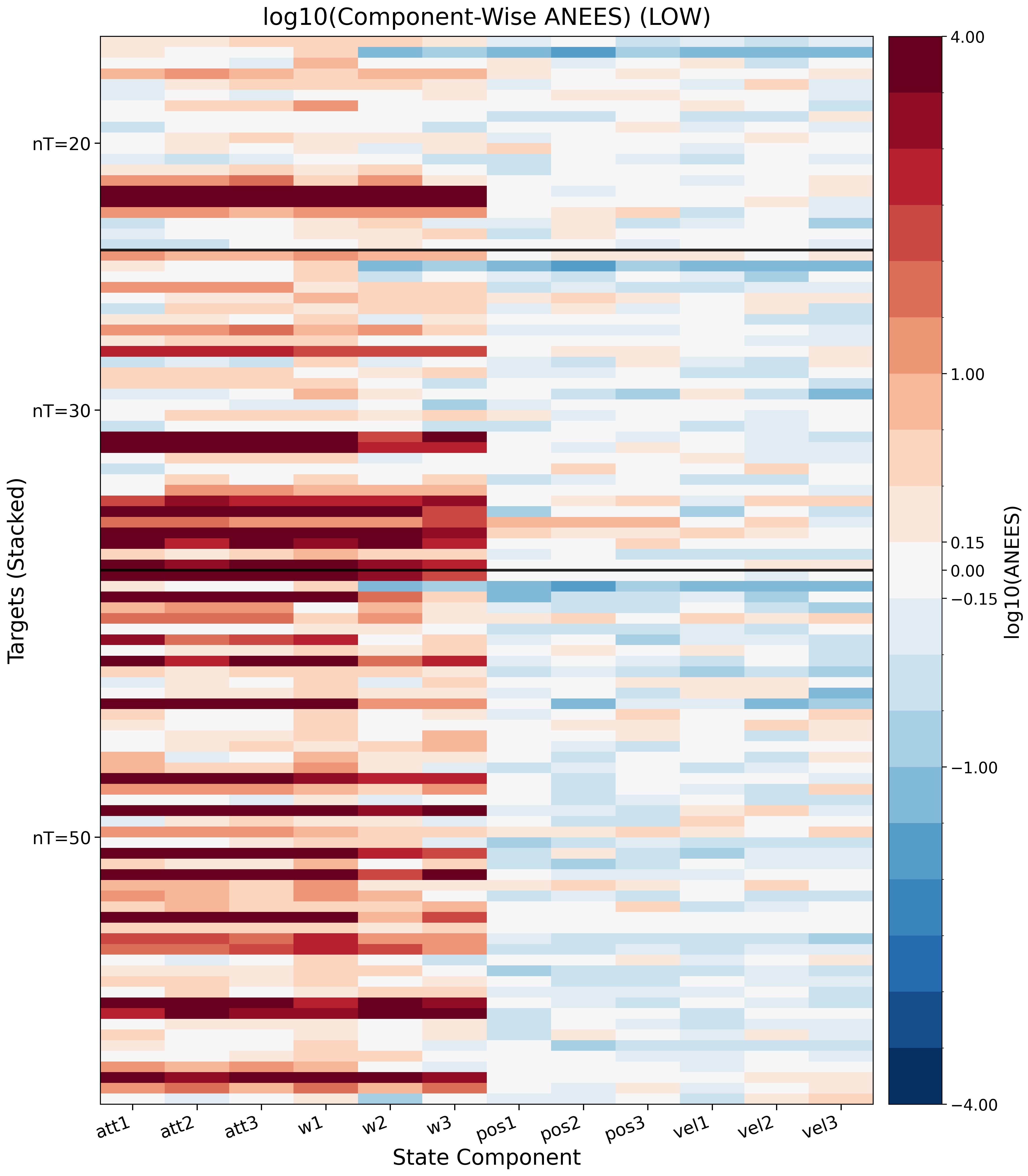}
         \caption{Component-Wise ANEES ($n_T=20,30,50$)}
         \label{allowedtargetstest_scA_3}
     \end{subfigure}
     \begin{subfigure}[h]{0.49\linewidth}
         \centering
         \includegraphics[width=.7\linewidth]{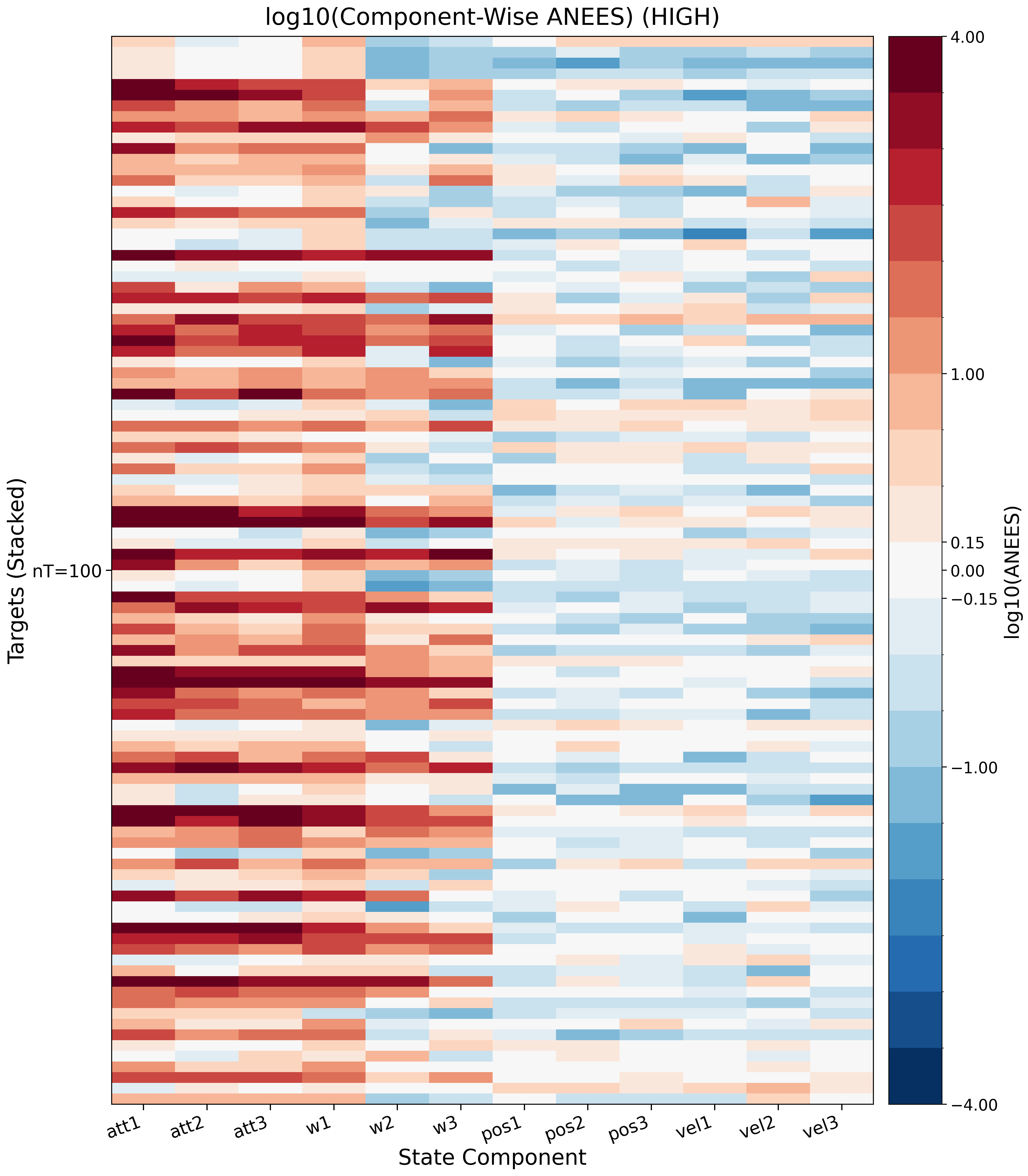}
         \caption{Component-Wise ANEES ($n_T=100$)}
         \label{allowedtargetstest_scA_4}
     \end{subfigure}       
     \begin{subfigure}[h]{0.49\linewidth}
         \centering
         \includegraphics[width=.7\linewidth]{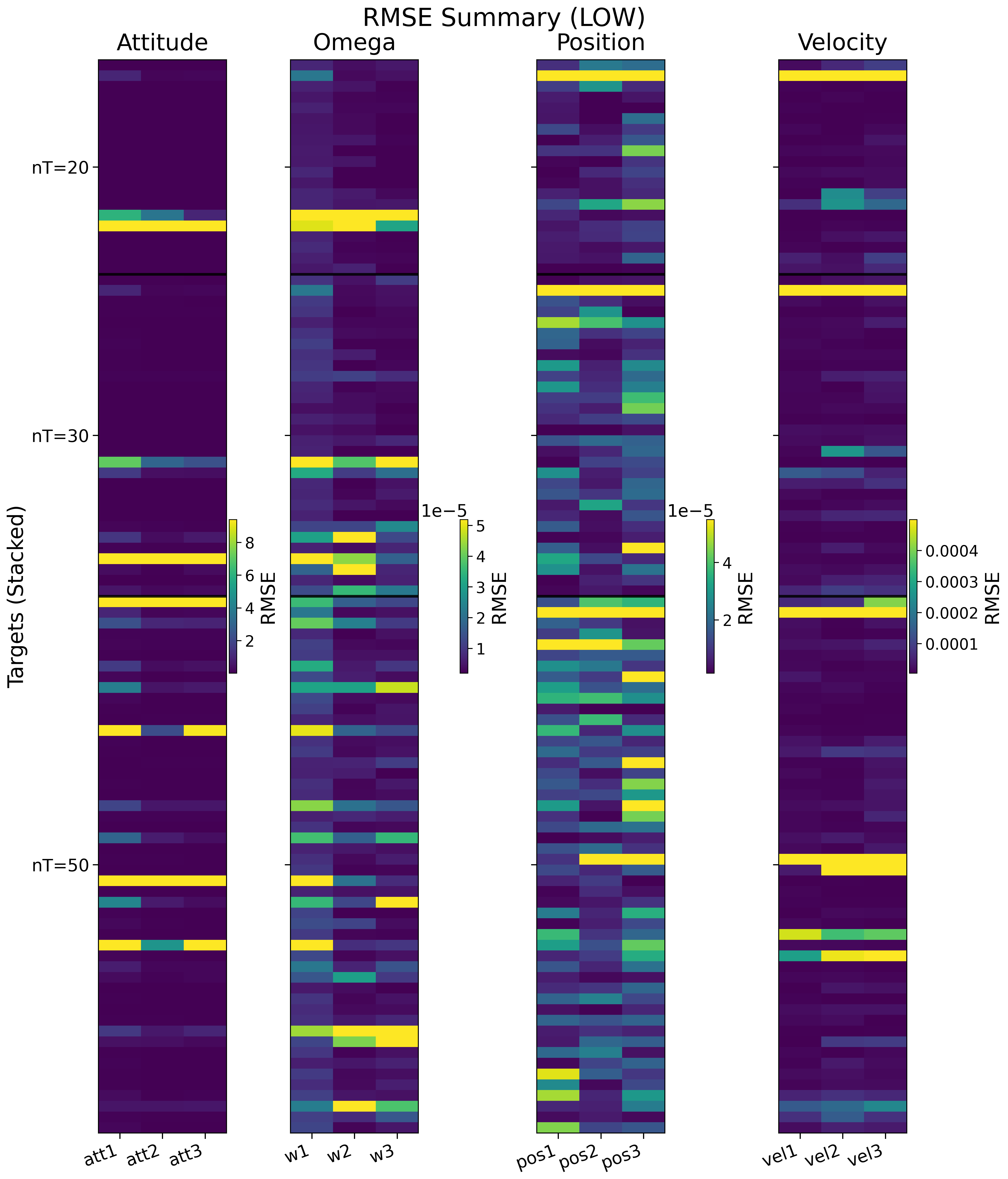}
         \caption{RMSE ($n_T=20,30,50$)}
         \label{allowedtargetstest_scA_5}
     \end{subfigure}
     \begin{subfigure}[h]{0.49\linewidth}
         \centering
         \includegraphics[width=.7\linewidth]{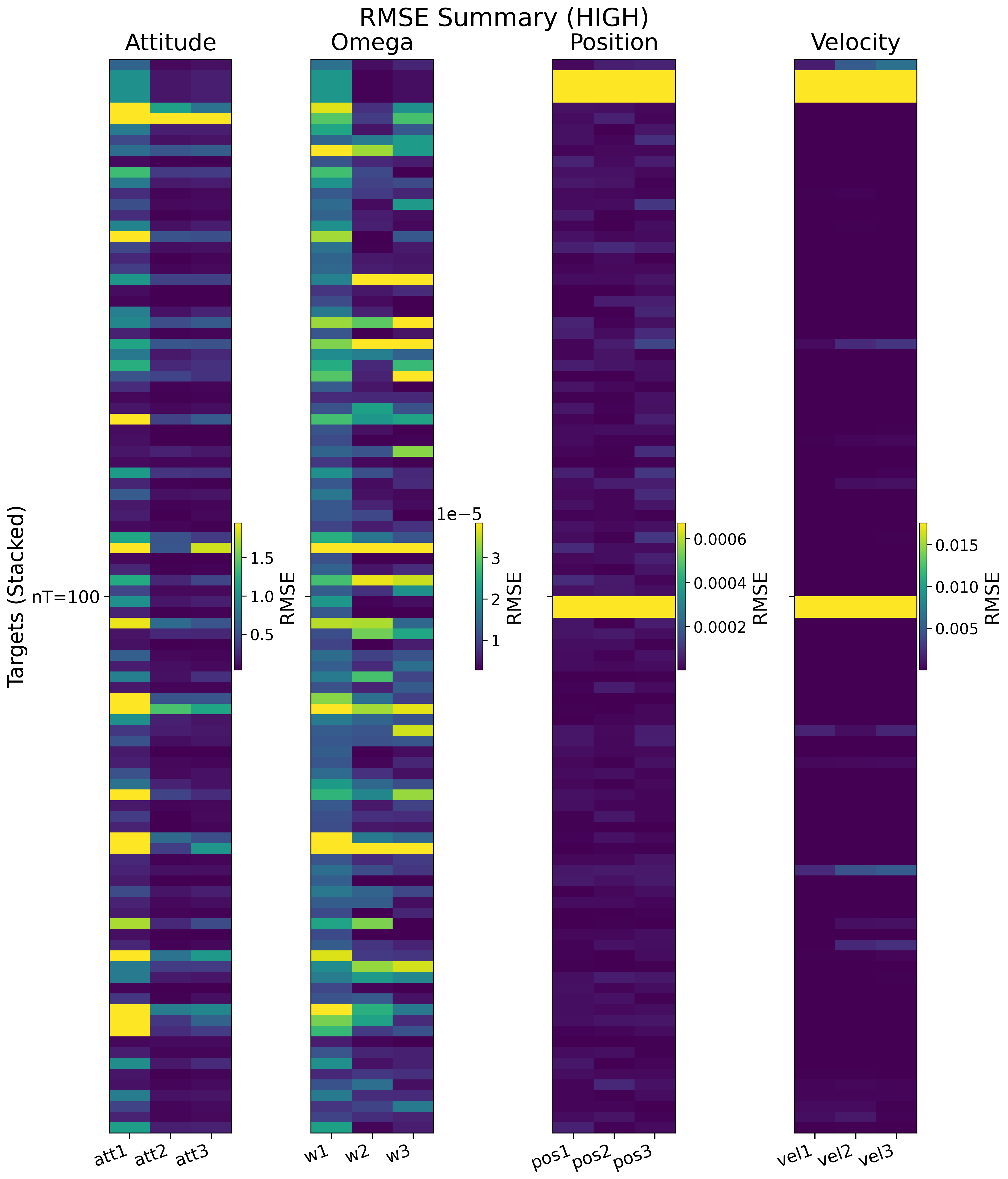}
         \caption{RMSE ($n_T=100$)}
         \label{allowedtargetstest_scA_6}
     \end{subfigure}         
        \caption{Scenario A-Based Estimation Performance for Different Target Counts from the Dual Sensor Tasking-Estimation Framework. (a), (b) Show the Fraction of Time Estimation Error is Within $3\sigma$ Bounds. (c), (d) Show Component-Wise ANEES. (e), (f) Show RMSE Values.}
        \label{allowedtargetstest_scA}
\end{figure}

\begin{figure}
    \centering
    \includegraphics[width=\linewidth]{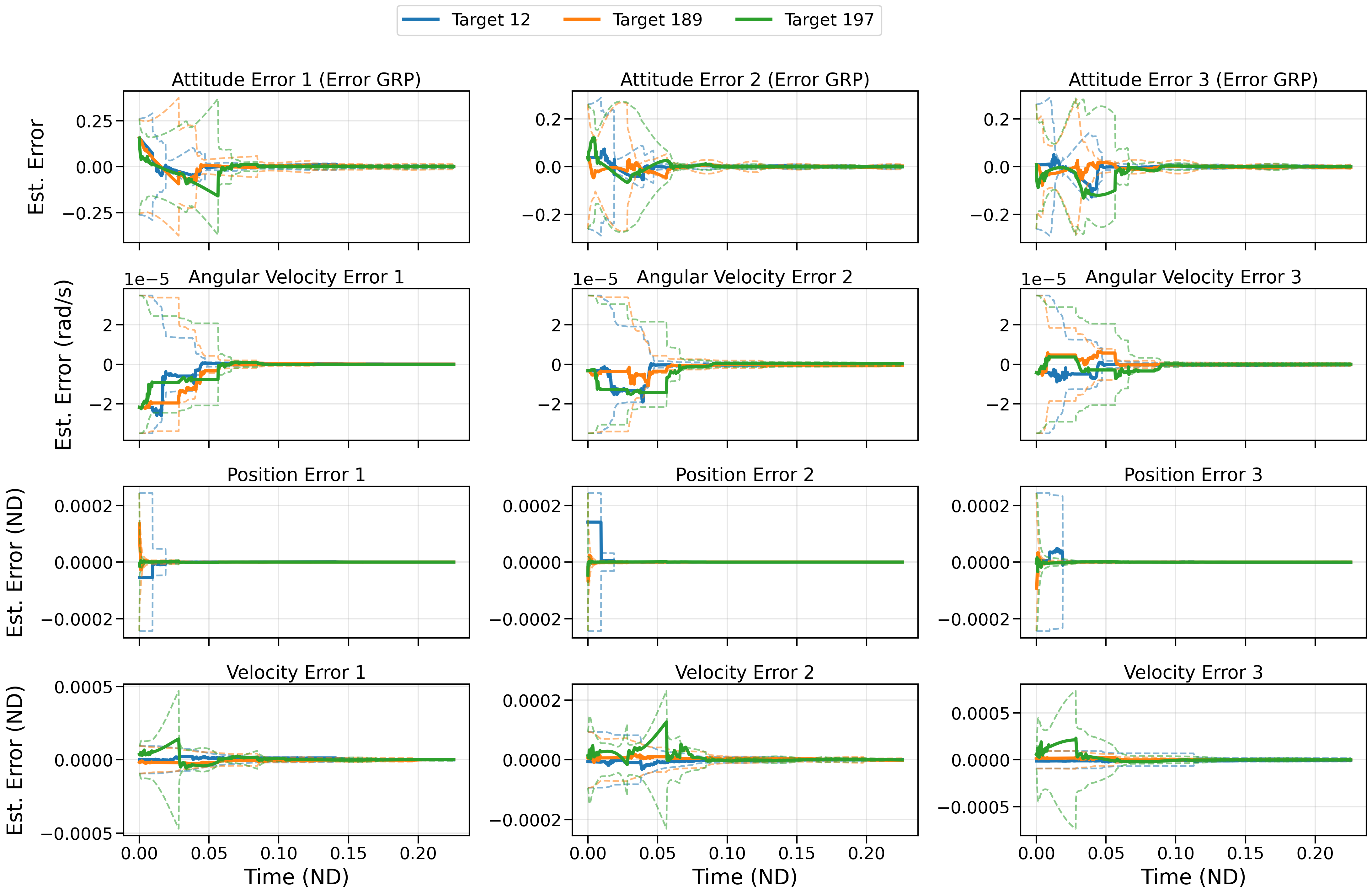}
    \caption{Estimation Results for a Select Few Targets Corresponding to $n_T=20$ (for Scenario A-based Architecture).}
    \label{selectfew_nt20_sca}
\end{figure}

Figure~\ref{allowedtargetstest_scA} shows the estimation performance from the dual sensor tasking-estimation framework utilizing the Scenario A-based architecture (20 observers) for different target counts. For brevity, only results corresponding to $n_T=20, 30, 50$ (relatively low target counts) and $100$ (relatively high target count) are shown. In Figs. \ref{allowedtargetstest_scA_1} and \ref{allowedtargetstest_scA_2}, we present the fraction of time during which the estimation errors remain within $3\sigma$ bounds for all the state components. Each colorized row corresponds to an individual target object. In Fig. \ref{allowedtargetstest_scA_1}, the separation between blocks (demarcated by the black solid line) highlights results for different $n_T$ cases. Figures~\ref{allowedtargetstest_scA_3} and \ref{allowedtargetstest_scA_4} show logarithm of component-wise average normalized estimation error squared (ANEES). For a given state component $i$, the ANEES is defined as:
\begin{equation}
    ANEES_i = \frac{1}{K} \sum_{k=1}^{K}\frac{e_i(k)^2}{\sigma_i(k)^2}
\end{equation}
where $e_i(k)$ and $\sigma_i(k)$ denote the estimation error and the associated standard deviation of component $i$ at time step $k$, respectively. ANEES values close to 1 (or log(ANEES) values close to 0) indicate statistically consistent estimation, values significantly greater than unity indicate an over-confident filter, while values significantly less than unity indicate an under-confident filter. Finally, Figs.~\ref{allowedtargetstest_scA_5} and \ref{allowedtargetstest_scA_6} show root mean square error (RMSE) values. In the RMSE plots, the attitude RMSE values are non-dimensional because they are calculated from error GRPs. Angular velocity RMSE values retain physical units of $rad/s$, whereas the position and velocity RMSE values are non-dimensional (w.r.t the characteristic length and time scales of the CR3BP rotating frame). A collective assessment of the three performance metrics in Fig.~\ref{allowedtargetstest_scA}, along with visual inspection of the estimation error and $3\sigma$ bounds for all targets (omitted for brevity), reveal following notable insights: (1) for the case $n_T=20$ (number of monitored objects is equal to the number of observers), the sensor tasking-estimation framework maintains satisfactory performance across both rotational and translational states for most targets. Estimation results corresponding to a select few targets are shown in Fig.~\ref{selectfew_nt20_sca}. Of the 20 targets considered, one is not observed throughout the simulation horizon, and two shows divergence in the rotational state estimates, even though their translational state estimates converge; (2) as the number of monitored cislunar targets increases relative to the number of cislunar observers, the estimation of rotational states across the target set becomes progressively more difficult, whereas the translational state estimates continue to exhibit satisfactory performance. The attitude estimation performance degradation can be attributed to a combination of insufficient attitude observability, fewer informative observations per target, and increased sensitivity of the filter to tuning parameters under sparse or unfavorable observation conditions. Although omitted for brevity, similar estimation trends are also observed in cases of Scenario B-based (33 observers) and C-based (43 observers) architectures.

\subsubsection{Sensor Tasking-Estimation Performance: Varying the Time Between Tasking Steps}
\begin{figure}[htbp]
     \centering
     \begin{subfigure}[h]{0.49\linewidth}
         \centering
         \includegraphics[width=.7\linewidth]{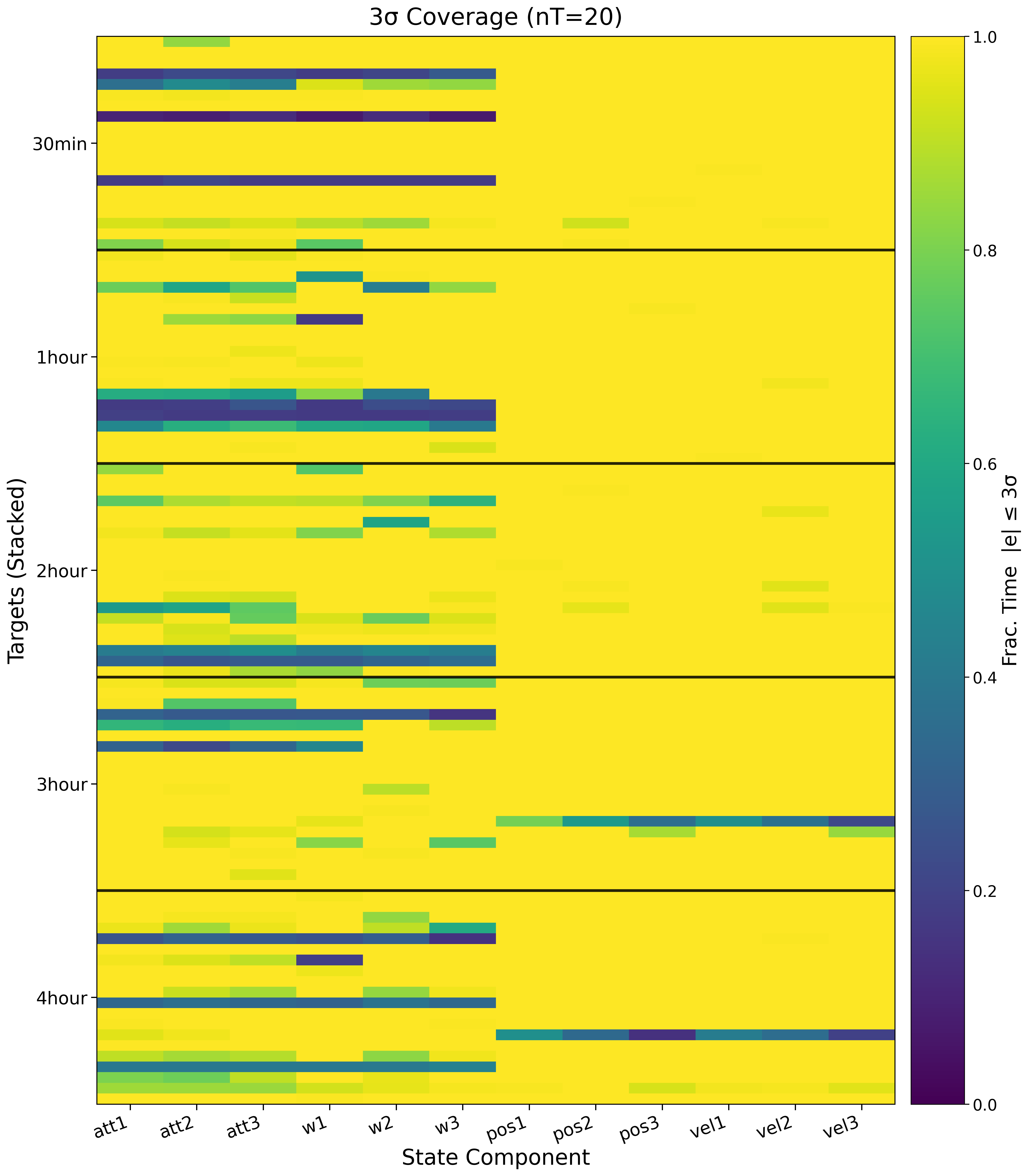}
         \caption{Fraction of Time $|e|\leq 3\sigma$ (Scenario-A-Based Architecture ($n_T=20$))}
         \label{deltasttest_1}
     \end{subfigure}
     \begin{subfigure}[h]{0.49\linewidth}
         \centering
         \includegraphics[width=.7\linewidth]{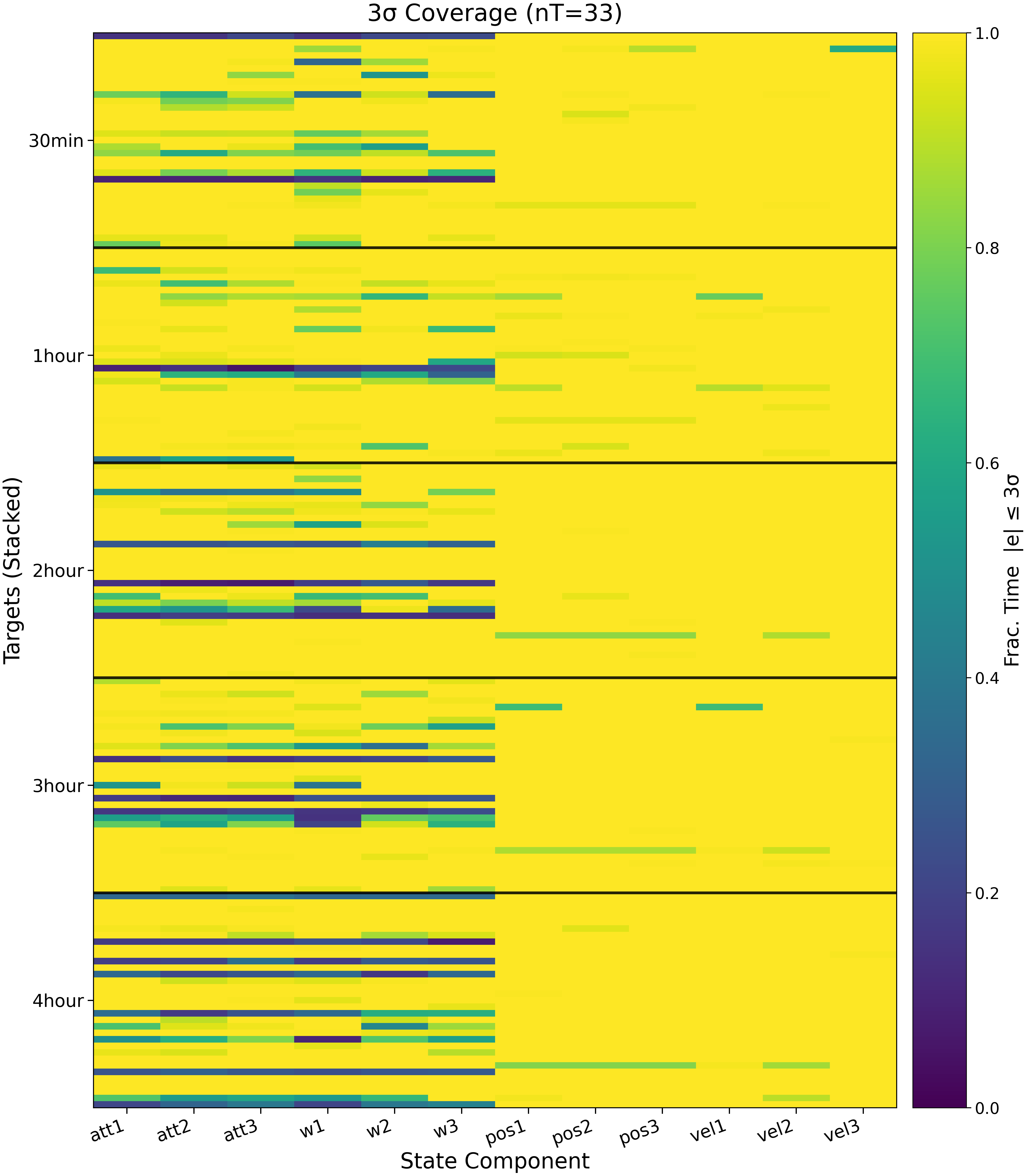}
         \caption{Fraction of Time $|e|\leq 3\sigma$ (Scenario-B-Based Architecture ($n_T=33$))}
         \label{deltasttest_2}
     \end{subfigure}   
     \begin{subfigure}[h]{0.49\linewidth}
         \centering
         \includegraphics[width=.7\linewidth]{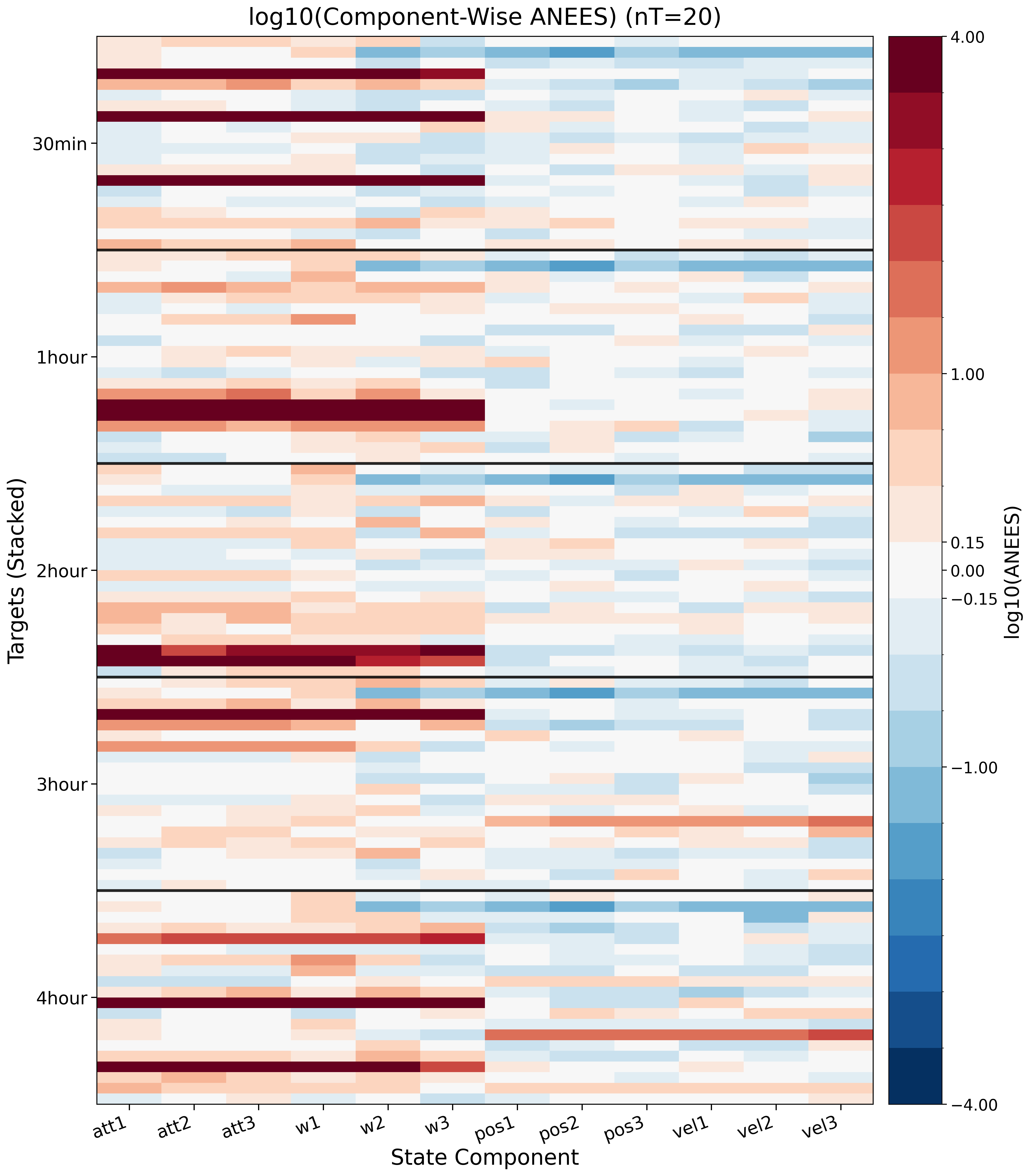}
         \caption{Component-Wise ANEES (Scenario-A-Based Architecture ($n_T=20$))}
         \label{deltasttest_3}
     \end{subfigure}
     \begin{subfigure}[h]{0.49\linewidth}
         \centering
         \includegraphics[width=.7\linewidth]{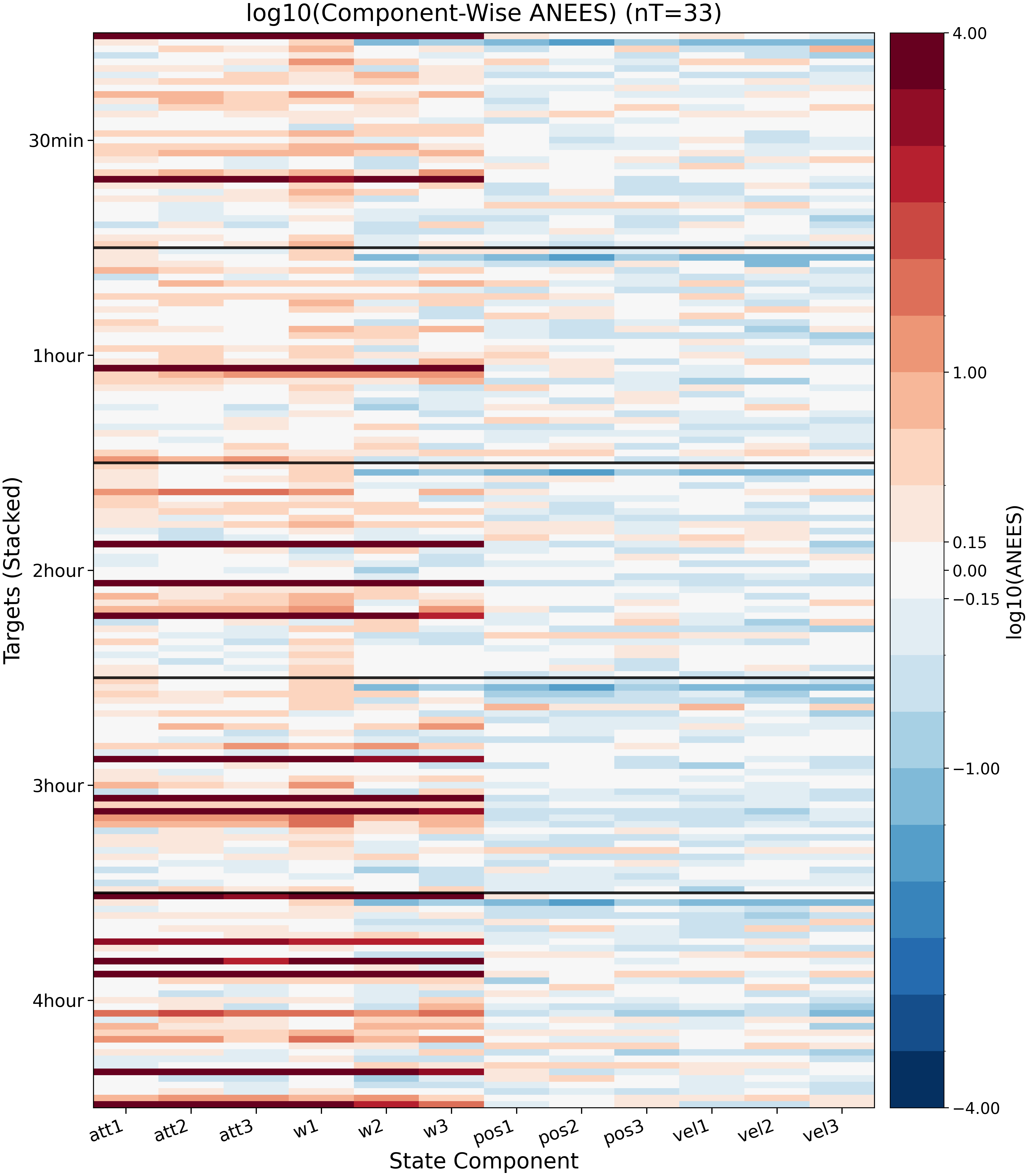}
         \caption{Component-Wise ANEES (Scenario-B-Based Architecture ($n_T=33$))}
         \label{deltasttest_4}
     \end{subfigure}       
     \begin{subfigure}[h]{0.49\linewidth}
         \centering
         \includegraphics[width=.7\linewidth]{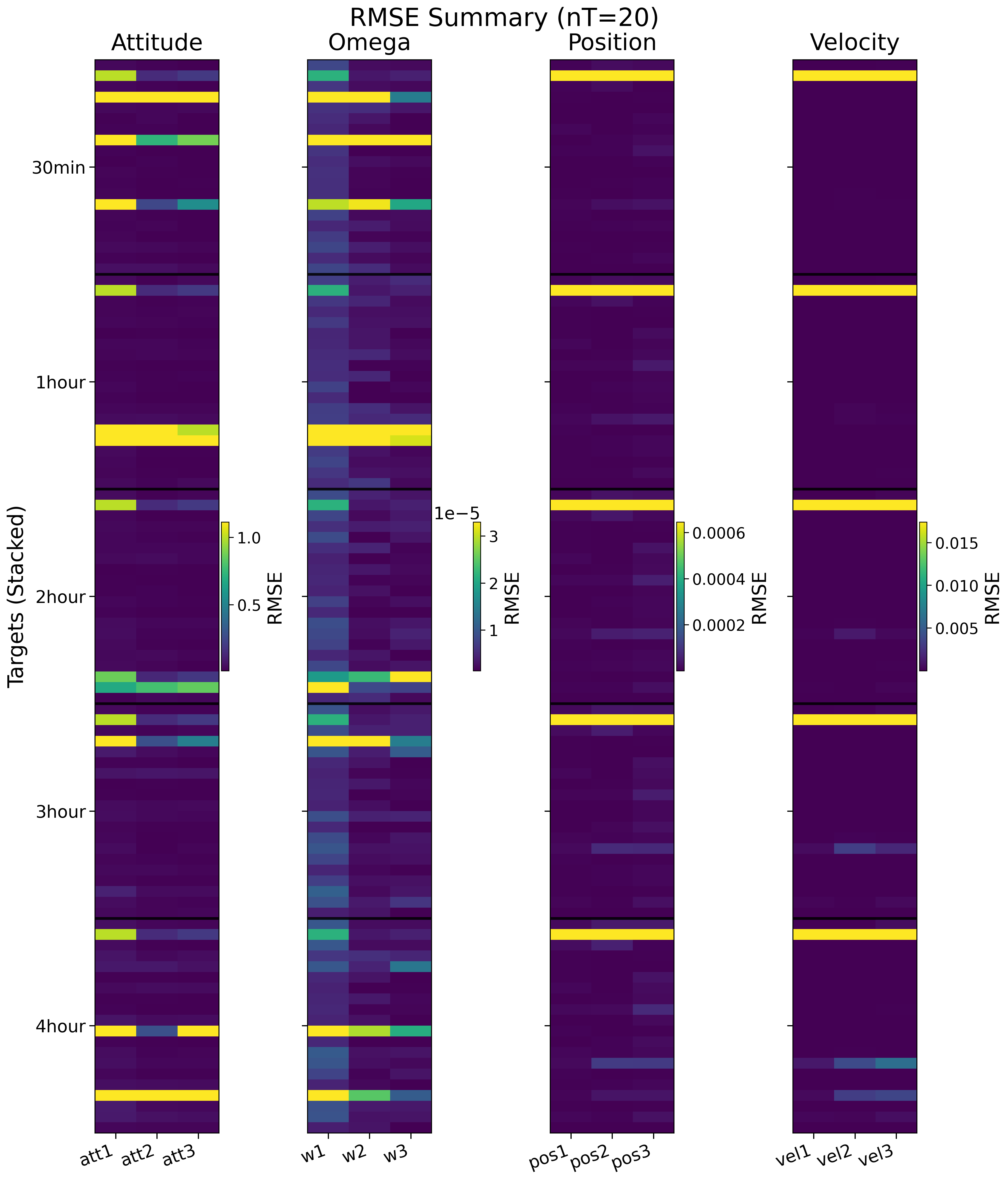}
         \caption{RMSE (Scenario-A-Based Architecture ($n_T=20$))}
         \label{deltasttest_5}
     \end{subfigure}
     \begin{subfigure}[h]{0.49\linewidth}
         \centering
         \includegraphics[width=.7\linewidth]{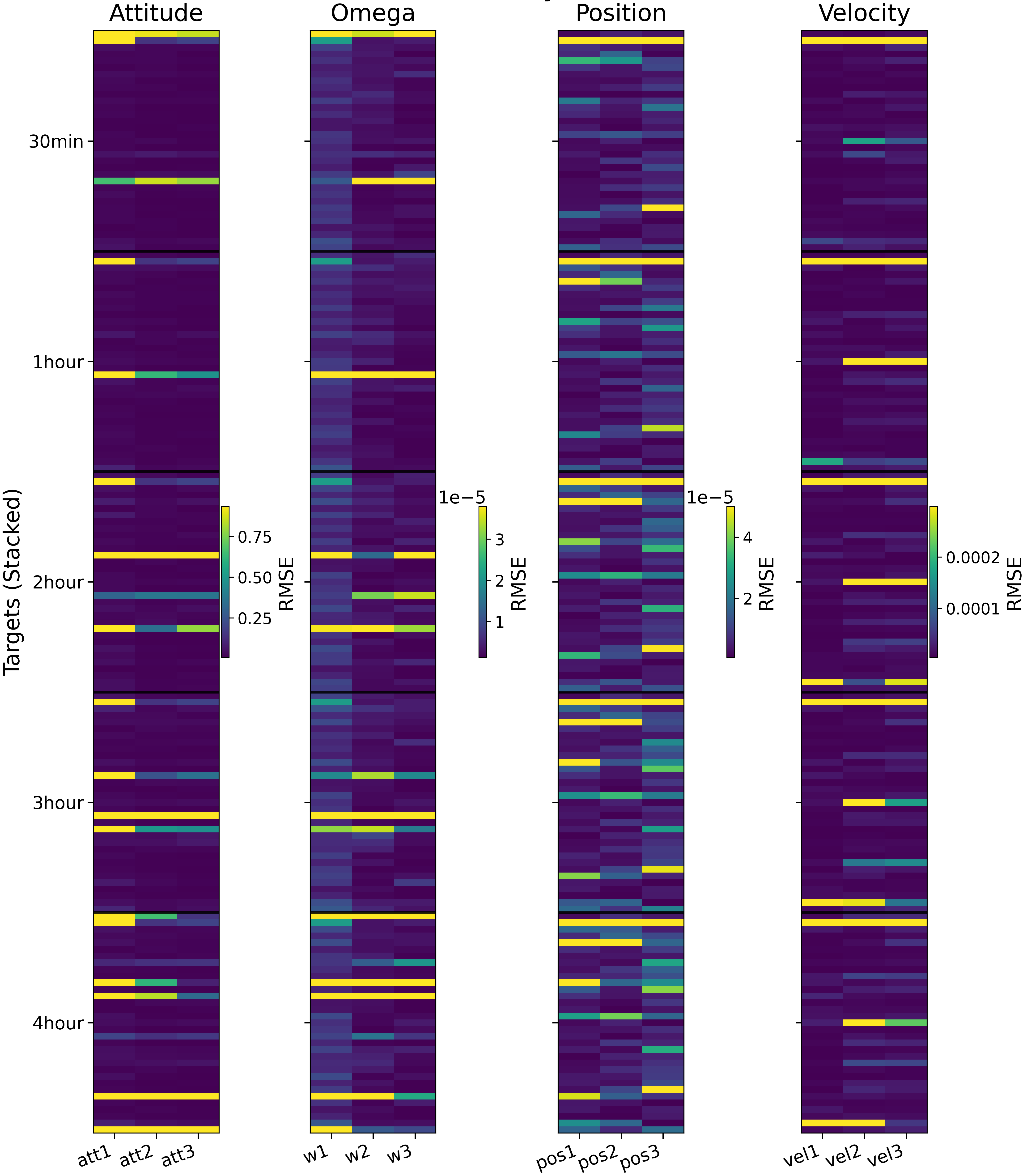}
         \caption{RMSE (Scenario-B-Based Architecture ($n_T=33$))}
         \label{deltasttest_6}
     \end{subfigure}         
        \caption{Scenario A and B-Based Estimation Performance for Different Tasking Intervals from the Dual Sensor Tasking-Estimation Framework. (a), (b) Show the Fraction of Time Estimation Error is Within $3\sigma$ Bounds. (c), (d) Show Component-Wise ANEES. (e), (f) Show RMSE Values.}
        \label{deltasttest}
\end{figure}

In this investigation, we analyze how varying the time interval between consecutive sensor tasking optimization steps impacts the estimation performance of orbital and rotational states. For brevity, we first present results for two representative cases with similar observer-to-target ratios: Scenario A-based architecture with 20 observers monitoring 20 targets and Scenario B-based architecture with 33 observers monitoring 33 targets. The following sensor tasking intervals are examined: $T_{ST}=\{30\;min,1H,2H,$ $3H,4H\}$. Figure~\ref{deltasttest} shows the estimation performance from the dual sensor tasking-estimation framework for different $T_{ST}$. Figures~\ref{deltasttest_1} and \ref{deltasttest_2} show the fraction of time for which the estimation errors remain within the $3\sigma$ bounds across all state components for the Scenario-A and Scenario-B-based cases, respectively. The logarithm of component-wise ANEES is shown in Figs.~\ref{deltasttest_3} and ~\ref{deltasttest_4}. Lastly, Figs.~\ref{deltasttest_5} and \ref{deltasttest_6} present the RMSE values. For the Scenario-A-based $T_{ST}$ test, the translational state estimates show only a modest degradation as the sensor tasking interval increases from $T_{ST}=30\;min$ to $T_{ST}=4H$, although the translational estimation accuracy and consistency remain satisfactory across the full range of tested $T_{ST}$. The rotational state estimates for Scenario-A do not exhibit a clear trend with respect to the sensor tasking interval across the tested $T_{ST}$ values, although the component-wise ANEES indicates a modest improvement at $T_{ST}=3H$. For the Scenario-B-based architecture, the rotational state estimates show a noticeable degradation with increasing sensor tasking interval, as evidenced by reduced $3\sigma$ coverage and elevated component-wise ANEES and RMSE between $T_{ST}=30\;min$ and $T_{ST}=4H$. Visual inspection of the estimation errors and associated $3\sigma$ bounds for all targets (omitted for brevity) suggests an increased occurrence of attitude estimation divergence for $T_{ST}=4\,H$ relative to $T_{ST}=30\;min$. The translational state estimates, by comparison, show only a modest degradation (performance still satisfactory) as the sensor tasking interval increases from $T_{ST}=30\;min$ to $T_{ST}=4H$.

\begin{figure}[htbp]
     \centering
     \begin{subfigure}[h]{0.49\linewidth}
         \centering
         \includegraphics[width=.7\linewidth]{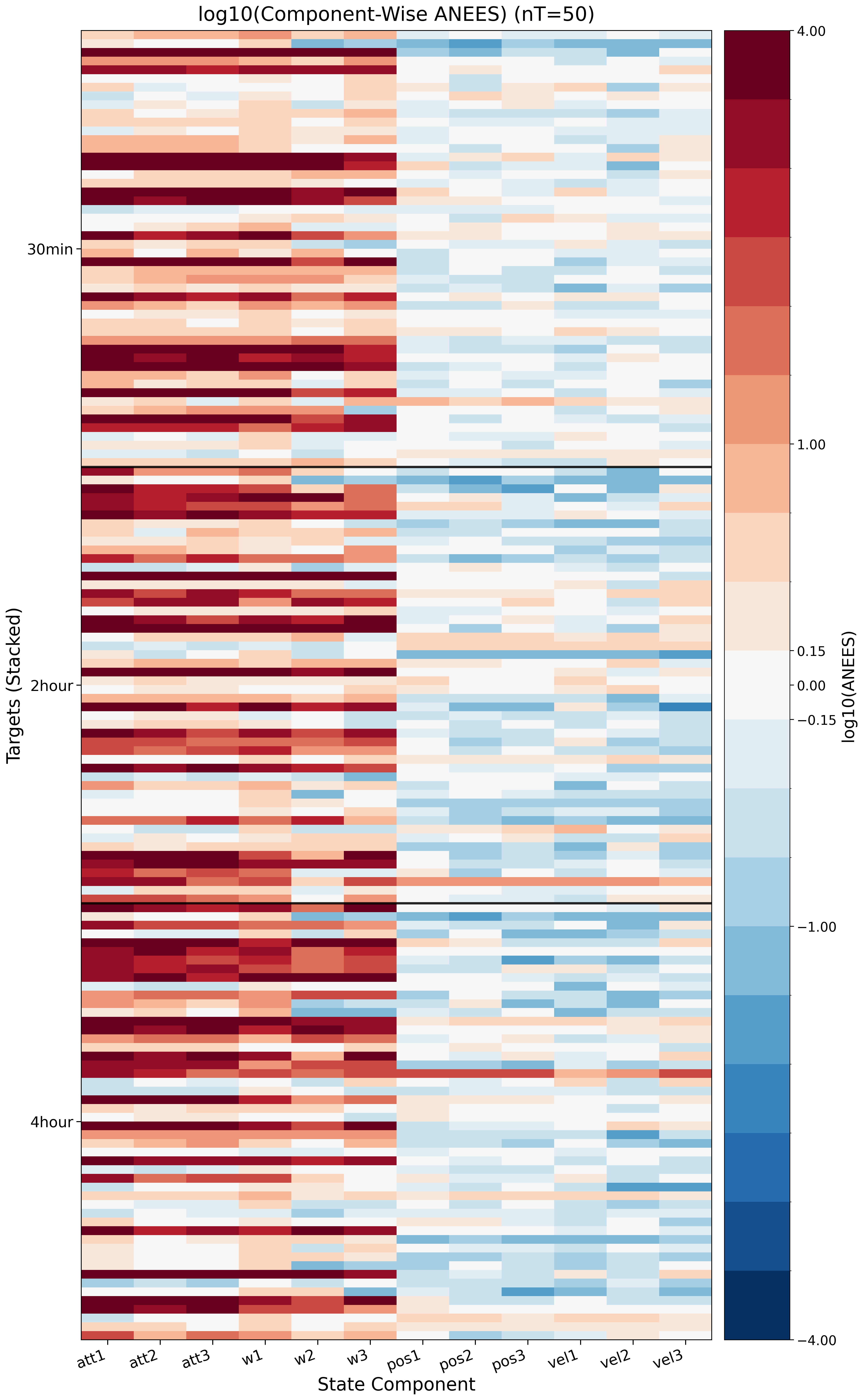}
         \caption{Component-Wise ANEES (Scenario-A-Based Architecture ($n_T=50$))}
         \label{deltasttestdiffload_1}
     \end{subfigure}
     \begin{subfigure}[h]{0.49\linewidth}
         \centering
         \includegraphics[width=.7\linewidth]{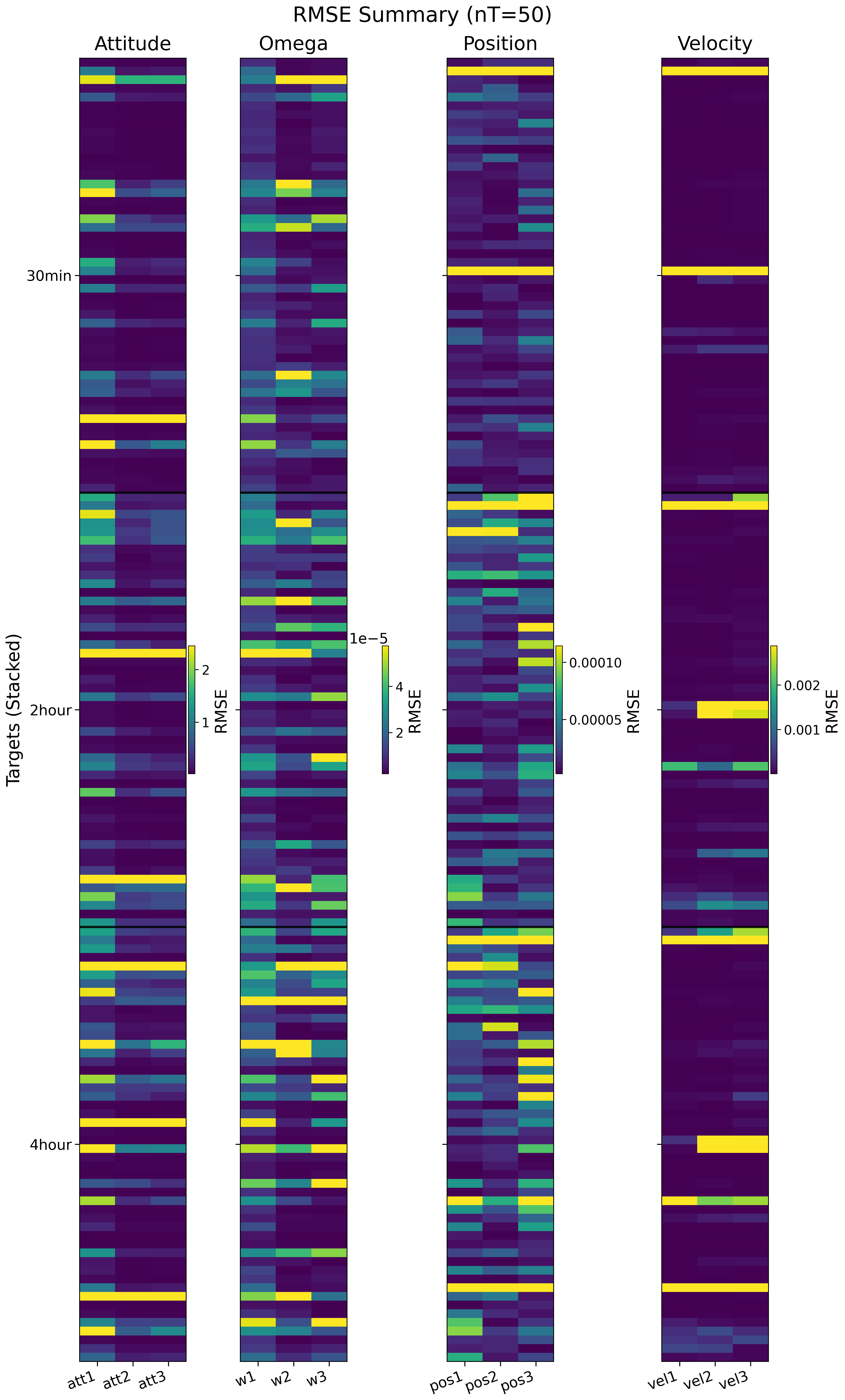}
         \caption{RMSE (Scenario-A-Based Architecture ($n_T=50$))}
         \label{deltasttestdiffload_2}
     \end{subfigure}   
     \begin{subfigure}[h]{\linewidth}
         \centering
         \includegraphics[width=\linewidth]{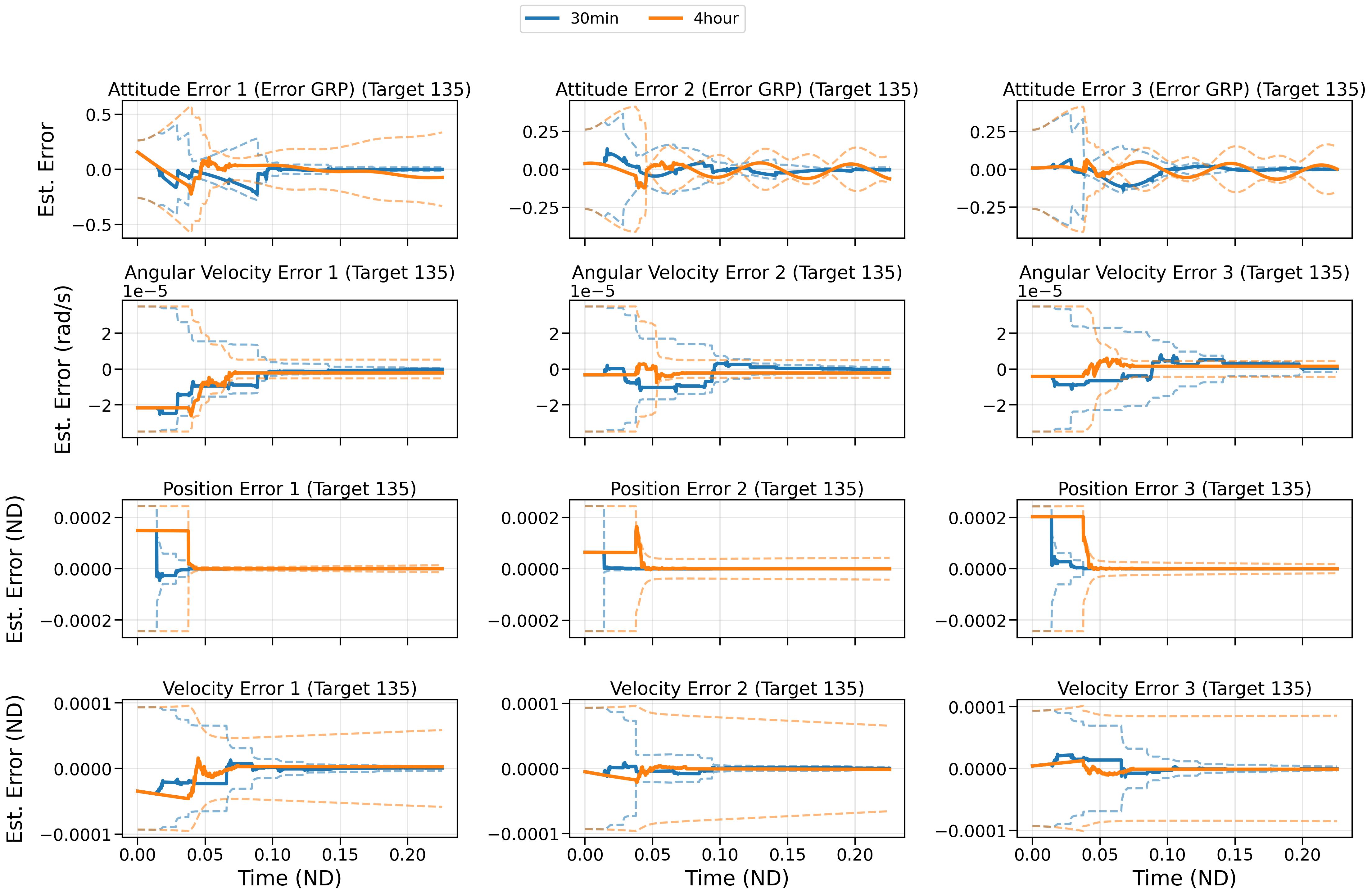}
         \caption{Estimation Performance for a Target ($T_{ST}=30\;min\;Vs\;4H$)}
         \label{deltasttestdiffload_3}
     \end{subfigure}
        \caption{Scenario A-Based Estimation Performance for Different Tasking Intervals from the Dual Sensor Tasking-Estimation Framework for a Higher Target-to-Observer Ratio (20 Observers, 50 Targets). (a) Shows Component-Wise ANEES, (b) Shows RMSE Values, and (c) Shows Estimation Results for a Representative Target.}
        \label{deltasttestdiffload}
\end{figure}

Next, we present results for a higher observer-to-target ratio corresponding to a Scenario A-based architecture with 20 observers monitoring 50 targets. Figure~\ref{deltasttestdiffload} shows the estimation performance from the dual sensor tasking-estimation framework for different $T_{ST}$ values. The logarithm of component-wise ANEES is shown in Fig.~\ref{deltasttestdiffload_1} and the RMSE values are shown in Fig.~\ref{deltasttestdiffload_2}. The degradation of estimation performance becomes increasingly pronounced as we move from $T_{ST}=30\;min$ to $T_{ST}=4H$. Multiple divergence events are observed in the rotational states, while the translational states remain comparatively well behaved. Estimation error curves along with $3\sigma$ bounds are shown in Fig.~\ref{deltasttestdiffload_3} for a representative non-diverging target, illustrating the deterioration observed for $T_{ST}=4H$ compared to $T_{ST}=30\;min$.

\section{Conclusion}\label{sec:conclusions}
This paper presents a unified framework for the design and operation of a space-based optical sensor network for cislunar space domain awareness. A novel composite cost function is proposed to determine the optimal observer architecture by jointly balancing satellite count, periodic orbit's stability, cumulative observability, per-target observation count, and proximity to the primary bodies. The composite cost is evaluated over a representative set of static targets defined in the rotating CR3BP frame, obtained by sampling mission-relevant zero-velocity surfaces and consolidating the resulting points through $k$-means clustering. Building on this optimal architecture, a mutual information-based sensor tasking strategy is introduced. The effectiveness of the sensor tasking is reflected in the resulting state estimation performance, demonstrated through joint translational and attitude estimation in a multi-object tracking setting, where sensor-target assignments are updated at fixed tasking intervals while estimation proceeds at a finer temporal cadence between updates.

The architecture design results in constellations ranging from 20 to 43 observers across the three static target-count scenarios in Task. Resulting observer orbits spread over ten selected CR3BP periodic orbits, with $L_2$ Halo north and $L_2$ Halo south orbit family being the more preferred observer orbits. For Task 2, it is not surprising to see satisfactory translational and rotational states estimation when there are equal number of observers and dynamically propagated targets. As the number of monitored targets increases relative to the number of observers, rotational state estimation becomes progressively more challenging than translational state, leading to more frequent divergence events. When the time between tasking steps is increased from 30 minutes to 4 hours, translational estimation degrades only modestly, whereas rotational estimation is more sensitive, exhibiting increased uncertainty and more frequent divergence, particularly at higher target-to-observer ratios.

While translational state is consistently estimated with high accuracy, attitude estimation performance remains comparatively degraded. This behavior is attributable to the reduced attitude observability of passive optical measurements under constrained viewing geometries, whereas translational states are more strongly informed by line-of-sight angular measurements. In this study, the mutual information metric used for tasking is computed from the orbital-state covariance and therefore does not explicitly prioritize rotational uncertainty within the tasking cost function. 

Future work will extend the present study by incorporating additional practical considerations, including heterogeneous sensing architectures, sensor fusion strategies, and inter-constellation communication constraints. Constellation geometries should also be examined with explicit emphasis on improving attitude observability, alongside other mission-level factors such as deployment and servicing costs. Finally, tasking strategies that more explicitly balance translational and rotational information gain represent a promising direction for further investigation.


\section*{Acknowledgments}
The authors gratefully acknowledge financial support from Missouri University of Science and Technology (Missouri S\&T) through institutional seed funding for new faculty and Embry-Riddle Aeronautical University (ERAU) through the Faculty Innovative Research in Science and Technology (FIRST) grant. The authors also thank Prof. John Crassidis for valuable insights and thoughtful discussions on the theoretical aspects of this work, Dr. Yang Cheng for review and feedback on the implementation, and Igor Panfil and Roei Nir for assistance in curating lunar mission orbits and reflection model implementations.

\bibliographystyle{jasr-model5-names}
\biboptions{authoryear}
\bibliography{references}

\end{document}